\documentclass{article}
\usepackage{graphicx} 
\usepackage{comment}
\usepackage[top=3cm,bottom=3cm,left=2cm,right=2cm,asymmetric]{geometry}
\usepackage{authblk}
\usepackage{hyperref} 
\usepackage{booktabs}
\usepackage{multirow}
\usepackage{textcomp}
\usepackage{gensymb}
\usepackage{tabularx}
\usepackage{caption}
\usepackage{footnote}
\newcolumntype{M}[1]{>{\centering\arraybackslash}m{#1}}
\usepackage{xcolor}
\usepackage{amssymb}
\usepackage{multicol}
\usepackage{enumitem}
\usepackage{amsmath}
\usepackage[ruled,vlined]{algorithm2e}
\usepackage{graphicx} 
\usepackage{float} 

\SetKwComment{Comment}{\#~}{}
\SetStartEndCondition{ }{}{}%
\SetKwProg{Fn}{def}{\string:}{}
\SetKwFunction{Range}{range}
\SetKw{KwTo}{in}\SetKwFor{For}{for}{\string:}{}%
\SetKwIF{If}{ElseIf}{Else}{if}{:}{elif}{else:}{}%
\SetKwFor{While}{while}{:}{fintq}%
\AlgoDontDisplayBlockMarkers\SetAlgoNoEnd\SetAlgoNoLine

\date{}

\title{Wandering around: A bioinspired approach to visual attention through object motion sensitivity}
\author[1,*]{Giulia D'Angelo}
\author[2]{Victoria Clerico}
\author[3]{Chiara Bartolozzi}
\author[1]{Matej Hoffmann}
\author[4]{P. Michael Furlong}
\author[5,6]{Alexander Hadjiivanov}
\affil[1]{Department of Cybernetics, Faculty of Electrical Engineering, Czech Technical University in Prague, Czech Republic}
\affil[2]{IBM Research Europe, Zurich, Switzerland}
\affil[3]{Event-Driven Perception for Robotics, Italian Institute of Technology, Genoa, Italy}
\affil[4]{National Research Council of Canada \& Systems Design Engineering, University of Waterloo, Canada}
\affil[5]{Advanced Concepts Team, European Space Agency, Noordwijk, The Netherlands}
\affil[6]{Adapsent, Leiden, The Netherlands}

\affil[*]{Corresponding Author: giulia.dangelo@fel.cvut.cz}

\begin{document}
\maketitle

\begin{abstract}
Active vision enables dynamic and robust visual perception, offering an alternative to the static, passive nature of feedforward architectures commonly used in computer vision, which depend on large datasets and high computational resources. Biological selective attention mechanisms allow agents to focus on salient Regions of Interest (ROIs), reducing computational demand while maintaining real-time responsiveness. 

Event-based cameras, inspired by the mammalian retina, further enhance this capability by capturing asynchronous scene changes, enabling efficient, low-latency processing.
To distinguish moving objects while the event-based camera is also in motion, the agent requires an object motion segmentation mechanism to accurately detect targets and position them at the center of the visual field (fovea).
Integrating event-based sensors with neuromorphic algorithms represents a paradigm shift, using Spiking Neural Networks (SNNs) to parallelise computation and adapt to dynamic environments. This work presents a Spiking Convolutional Neural Network (sCNN) bioinspired attention system for selective attention through object motion sensitivity.
The system generates events via fixational eye movements using a Dynamic Vision Sensor (DVS) integrated into the Speck neuromorphic hardware, mounted on a Pan-Tilt unit, to identify the ROI and saccade toward it.

The system, characterised using ideal gratings and benchmarked against the Event Camera Motion Segmentation Dataset (EVIMO), reaches a mean IoU of 82.2\% and a mean SSIM of 96\% in multi-object motion segmentation. Additionally, the detection of salient objects reaches an accuracy of 88.8\% in office scenarios and 89.8\% in challenging indoor and outdoor low-light conditions, as evaluated on the Event-Assisted Low-Light Video Object Segmentation Dataset (LLE-VOS). 
A real-time demonstrator showcases the system's capabilities of detecting the salient object through object motion sensitivity in $0.124$ seconds in dynamic scenes.
Its learning-free design ensures robustness across diverse perceptual scenes, making it a reliable foundation for real-time robotic applications and serving as a basis for more complex architectures.
\end{abstract}

\begin{figure}[ht!]
    \centering
    \includegraphics[width=1\textwidth]{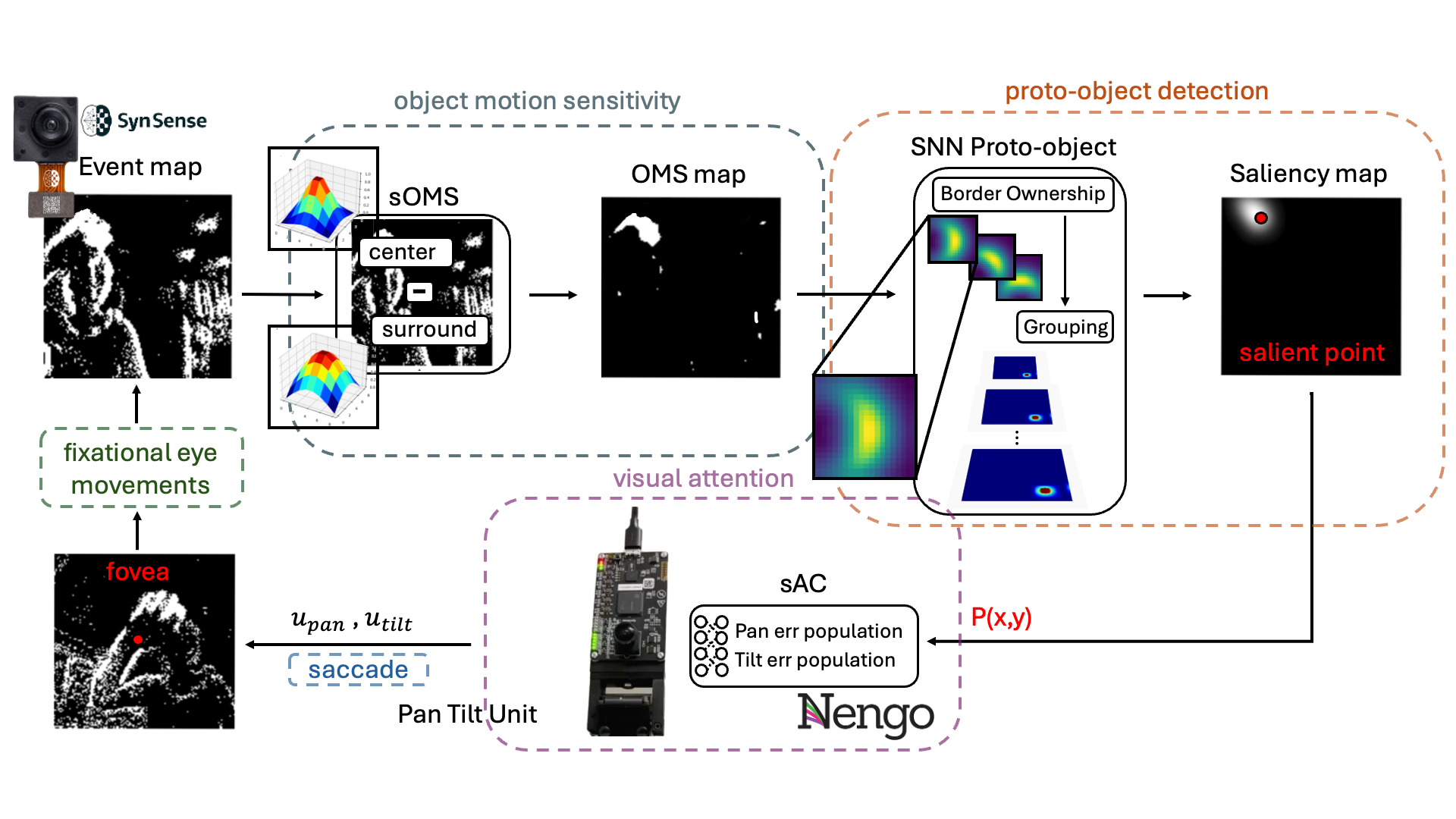}
    \caption{Overview of the system: From left to right, events from the Dynamic Vision Sensor (DVS) integrated into the Speck device are processed from left to right. These events enter the \textit{object motion sensitivity} module, where they are processed by the spiking Object Motion Sensitive (sOMS) model. This model generates the Object Motion Segmentation (OMS) map, which is then fed into the \textit{proto-object detection} module. The map is further processed by the Spiking Neural Network (SNN) Proto-object (SNN Proto-object) model, producing the final saliency map and identifying the most salient object (red circle, P($x,y$)). This triggers the Spiking Attention Control (sAC) mechanism in the \textit{visual attention} module, which generates pan and tilt control signals ($u_{pan}, u_{tilt}$) for the saccadic movement toward the salient object, using a FLIR Pan-Tilt Unit (PTU). The system then performs fixational eye movements to identify the next salient point and close the loop. See the accompanying video for a demonstration of the system.}
    \label{fig:pipeline}
\end{figure}

\textbf{Media}: The accompanying video can be found online\footnote{\href{https://youtu.be/dcAJlDgVR0o}{https://youtu.be/dcAJlDgVR0o}}.

\section{Introduction}

Active vision entails dynamically adjusting camera viewpoints to enhance environmental perception and information gathering. The concept gained prominence in the late 1980s through the work of Aloimonos \emph{et al.}~\cite{aloimonos1987active}, who demonstrated that an active observer can solve basic vision problems more effectively than a passive one. 
In contrast, passive perception has often been described as challenging because it fails to resolve ill-posed problems~\cite{hadamard1902problemes} (such as existence, uniqueness, and stability), lacking both a unique solution and constraints~\cite{poggio1984analog, aloimonos1987active}. This limitation results in ambiguities, making it difficult to determine a single, definitive interpretation of the visual data.
An active approach addresses the limitations of fixed-camera systems, such as occlusions and restricted fields of view.
By adapting dynamically to changing conditions, active vision enables a more comprehensive and unobstructed visual experience, something vital for navigating the physical world.
A key challenge in real-time robotics applications is balancing power consumption, latency, and data storage. Although traditional computer vision methods have achieved remarkable performance, these approaches struggle with generalisation, particularly in handling variations in scale, pose, and lighting conditions, or when identifying objects in complex or obstructed environments~\cite{zhu2020review}. Increasing data usage to overcome these challenges often exacerbates issues related to data processing, storage and transfer~\cite{feng2019computer}.

Given the agent's finite computational abilities, it is essential to allocate limited perceptual and cognitive resources. 
Attention mechanisms are frequently employed in robotics~\cite{azizevaluation,pasquale2016enabling,aarthi2020study} to selectively process Regions of Interest (ROIs), effectively reducing the volume of data to be processed and thereby facilitating real-time responsiveness. 
Visual attention in robotics offers numerous benefits, such as enhancing interaction efficiency by enabling robots to selectively focus on relevant stimuli consequently improving adaptability and responsiveness in dynamic environments~\cite{hanifi2024pipeline,potapova2017survey,pasquale2016enabling}. 

The process of filtering relevant information from the continuous bombardment of complex sensory data and shifting the centre of the visual field (fovea) toward the current ROI, mimics the biological mechanism of overt selective attention~\cite{rizzolatti1983mechanisms}.
This process occurs not only in animals, where selecting the most significant stimuli (such as the presence of a predator) is crucial but also in complex machinery with a diverse array of sensors, such as robots, where similar attentional mechanisms can be leveraged to focus the processing on the most salient part of the visual field. 
Models of bottom-up visual attention~\cite{itti1998model, itti2001computational, ude2005distributed} have been proposed to generate feature-based saliency maps for detecting salient regions or proto-objects~\cite{russell2014model}. 

To detect objects, the human visual perception of a scene strongly depends on perceptual organisation rules, known as the 'Gestalt laws'~\cite{kohler1967gestalt}, such as continuity, proximity, figure-ground organisation, and closure. 
These principles give perceptual saliency to regions of the visual field that can be perceived as ``proto-objects'', referring to areas of the scene that can potentially contain an object. To detect the presence of a proto-object, perceptual grouping mechanisms of the visual input refer to neuronal populations of the secondary visual cortex (V2)~\cite{zhou2000coding} encoding the geometric properties of a proto-object through their firing patterns. The sensitivity of these cells extends beyond edge orientation to encompass the concept of ``border ownership'', which defines the assignment of edges to the foreground or background, ultimately enabling the detection of proto-objects.

Several frame-based proto-object approaches have utilised a bioinspired architecture resembling border ownership cells~\cite{russell2014model} by applying the von Mises kernel with varying orientations and scales~\cite{todorovic2008gestalt} to identify areas of the visual field that potentially contain objects. These implementations feed different cues of information into the proto-object model, generating saliency maps dependent on the provided input, such as motion~\cite{molin2013proto}, depth~\cite{hu2016proto} and texture~\cite{uejima2020proto,uejima2018proto}.
These models have significantly influenced our understanding of vision and have inspired robotic implementations~\cite{orabona2005object}.

However, these models predominantly rely on traditional frame-based vision sensors, which are limited by restricted dynamic range, redundancy, motion blur, latency and high power consumption. 
Event-based cameras~\cite{Lichtsteiner2008}, which mimic the initial layers of the mammalian retina and react to pixel-level illumination changes, offer a solution to these limitations. Unlike frame-based cameras, event-based sensors improve dynamic range, reduce latency, and generate an asynchronous event stream providing information of spatial coordinates, polarity, and timestamps. This results in a significant reduction in data processing, making event-based cameras highly relevant for robotic applications~\cite{monforte2020exploiting,mueggler2018continuous,Iacono2018,glover2017robust}. Their inherent real-time response to luminance changes provides an ideal sensory input for guiding subsequent visual attention actions.
Previous implementations of event-driven proto-object models for the humanoid robot iCub~\cite{bartolozzi2011embedded}, leveraging cues of intensity~\cite{iacono2019proto} and depth~\cite{ghosh2022event}, have paved the way for fully bioinspired proto-object models in robotics, effectively bridging the gap between bioinspired algorithms and hardware.

While these models utilise the sparse, event-driven nature of visual sensors, their full potential is unrealised without execution on a neuromorphic platform. Most computer vision processing occurs on CPUs, GPUs, or specialized devices like Tensor Processing Units (TPUs) and Neural Processing Units (NPUs), which are optimized for dense data but require constant data transfer between the sensor, memory, and processor~\cite{MutluGhoseGomez-LunaEtAl_2019_ProcessingDataWhere}, consuming vast amounts of energy. Frame-based cameras also consume significant energy by capturing data continuously.
Neuromorphic computing~\cite{douglas1995neuromorphic} addresses these challenges by leveraging event-driven sensing and Spiking Neural Networks (SNNs) to process streams of events in parallel, inspired by the brain's efficiency~\cite{DiehlS2015SpikingDeepNetworks, 7838165,gehrig2020eventbasedangularvelocityregression,SpikeMS} demonstrating to be effective also for robotics real-time applications~\cite{fatahi2024event, Dangelo_etal22, d2020event}. Unlike von Neumann architectures, neuromorphic systems co-locate computation and memory, reducing data transfer and achieving greater energy efficiency. By mimicking biological neurons, neuromorphic circuits, whether analog~\cite{pehle2022brainscales,moradi2017scalable,qiao2015reconfigurable} or digital~\cite{Furber2014,davies2018loihi,sga2020neuromorphic}, overcome traditional processing limitations through SNNs and integrated memory and computation units.
A further advancement was the implementation of the proto-object model proposed by Iacono \emph{et al.}~\cite{iacono2019proto} on the SpiNNaker neuromorphic platform~\cite{Dangelo_etal22}. This effort~\cite{Dangelo_etal22} led to the development of the first end-to-end, saliency-based proto-object Spiking Neural Network (SNN), significantly reducing processing latency from $\sim$100 ms to approximately $\sim$16 ms.

All of the mentioned event-based and neuromorphic proto-object models implemented on the humanoid robot iCub~\cite{iacono2019proto,ghosh2022event,Dangelo_etal22} rely on stereotypical circular eye movements that mimic human fixational eye movements to perceive static scenes. The sparse event-based output from the sensors generates events not only in response to changes in the visual scene but also due to egomotion-induced motion of the camera. These egomotion-induced events overlay the relative motion of objects, requiring a mechanism to enhance the detection of objects' motion.
Object Motion Sensitivity (OMS) is critical to distinguishing between self-motion and objects' motion in the visual field, which, in turn, is essential for shifting attention towards moving items (such as prey and predators), and for effective navigation and interaction with the environment~\cite{schwartz2021object}.

Recent advancements in object motion sensitivity models have provided valuable insight into how biological visual systems enhance object motion during eye movements. {\"O}lveczky \emph{et al.}~\cite{olveczky2003segregation} and later Baccus \emph{et al.}~\cite{baccus2008retinal} investigated the neural circuitry underlying OMS in retinal ganglion cells, revealing their ability to selectively respond to relative motion between the center and the surrounding regions of the receptive fields. These cells suppress responses to global image motion, such as that generated by an observer's movements, enabling the identification of moving objects against dynamic backgrounds. These studies further pinpointed the involvement of specific retinal interneurons, including fast, transient bipolar cells and polyaxonal amacrine cells, which mediate the OMS response through carefully balanced excitatory and inhibitory pathways.

These cells have inspired recent advancements in modelling object motion sensitivity systems, deriving biologically motivated circuits~\cite{schwartz2021object}. The hardware implementation, Integrated Retinal Functionality in Image Sensors (IRIS)~\cite{yin2023iris, sinaga2024hardwarealgorithmreengineeringretinalcircuit} proposed a CMOS-compatible design of retinal-like computations, such as OMS, to enable low-power real-time in-sensor processing with minimal computational overhead. Further work by Clerico \emph{et al.}~\cite{clerico2024retina} proposed an algorithmic implementation of OMS showing state-of-the-art performance~\cite{parameshwara2021spikems,parameshwara20210,zhang2023multi} while utilising fewer parameters, making it particularly suitable for resource-constrained systems.
By processing DVS data, the algorithm achieves a significant reduction in computational complexity by a factor of 1000 compared to traditional methods~\cite{parameshwara2021spikems,parameshwara20210,zhang2023multi}, making it suitable for low-latency, low-bandwidth applications in dynamic environments.
The process employed by these algorithms to enhance relative motion does not ensure the identification of whether the motion results from a coherently moving object in the visual field or from a disjointed set of features that do not belong to a single object (\emph{i.e.}, clutter), thereby necessitating a mechanism for perceptual grouping such as the saliency-based proto-object model~\cite{Dangelo_etal22}.

The proposed work builds upon the visual attention work by D'Angelo \emph{et al.}\cite{Dangelo_etal22, ghosh2022event, iacono2019proto} and the object motion sensitivity model by Clerico \emph{et al.}\cite{clerico2024retina}, combining them to reduce the amount of information processed by focusing solely on the salient proto-object of the scene while enhancing object motion during camera motion.
In an effort to develop an architecture that significantly reduces power consumption and latency, this work bridges the gap between bio-inspired software and hardware by proposing a fully spiking Convolutional Neural Network (sCNN) architecture that leverages event-based sensors and neuromorphic algorithms.
Furthermore, the learning-free design of both algorithms ensures their applicability across diverse scenarios, demonstrating the system's capability to operate in new environments without requiring training datasets.

The online architecture, which enables the system to \textit{wander around}, enhance object motion detection through fixational eye movements, and perform saccades toward salient objects, is implemented using a Speck device mounted on a Pan-Tilt Unit (PTU). This setup explores the environment by processing events generated by the integrated DVS sensor on the Speck.
The Speck\footnote{\href{https://www.synsense.ai/products/speck-2/}{https://www.synsense.ai/products/speck-2/}} platform is the world’s first fully event-driven neuromorphic vision System-on-Chip (SoC) integrating a state-of-the-art DVS with a fully asynchronous chip architecture.
In the pursuit of fast online robotic applications, opting for the Speck neuromorphic hardware from the array of available hardware solutions emerges as the optimal choice due to its seamless integration with a DVS demonstrating low energy consumption~\cite{sorbaro2020optimizing} and suitability for common robotics online tasks~\cite{liu2019live}.
The events from the DVS sensor are processed through the Spiking Object Motion Sensitivity (sOMS) model to enhance motion segmentation and through the SNN Proto-object model to focus visual attention on a single maximum salient proto-object. The identified maximum salient point is subsequently fed into the Spiking Attention Control (sAC) model to move the PTU and position the maximum salient point at the center of the visual field. 
The system then executes small stabilisation movements, similar to fixational eye movements in the mammalian visual system~\cite{ko2010microsaccades}, to identify the next salient point and close the loop.
Such small and frequent movements are essential to prevent the rapid fading of static elements in the environment, which are crucial for building a saliency map for the next saccade.
An overview of the architecture is shown in Figure~\ref{fig:pipeline}. 

The contributions of this work can be summarised as follows: 
\begin{itemize}[noitemsep]
    \item Characterisation of the Spiking Object Motion Sensitivity CNN model (sOMS) through qualitative and quantitative assessments, inspired by the cell recordings conducted by {\"O}lveczky \emph{et al.}~\cite{olveczky2003segregation}, and an investigation of the parameters in the model proposed by Clerico \emph{et al.}~\cite{clerico2024retina}. 
    \item Characterisation of the SNN Proto-Object model investigating the model's parameters, as demonstrated in~\cite{iacono2019proto,ghosh2022event,Dangelo_etal22}, using the SalMapIROS dataset from~\cite{iacono2019proto}.
    \item Qualitative and Quantitative assessment of the sOMS model demonstrating the suppression of \textit{meaningless} events and benchmarking the model against the EVIMO dataset~\cite{mitrokhin2019ev} computing Intersection over Union (IoU) and Structural Similarity Index (SSIM) scores.
    \item Qualitative and quantitative benchmark to highlight the importance of incorporating the SNN Proto-object model for multi-object detection on the EVIMO dataset~\cite{mitrokhin2019ev}.
    \item Qualitative and quantitative benchmark in challenging low-light indoor and outdoor scenarios on the LLE-VOS dataset~\cite{li2024event} to demonstrate the system robustness.
    \item A real-time bioinspired attention system demonstrator showcasing the Speck device mounted on a PTU, wandering around and performing saccades toward the most salient object through \textit{object motion sensitivity} and fixational eye movements.
\end{itemize}

\section{Related work}\label{ch:relworks} 

Active vision enables intelligent systems to interact dynamically with their environment through mechanisms like selective visual attention, focusing on relevant details to achieve efficient and adaptive perception~\cite{rizzolatti1983mechanisms}. The human visual system is specifically structured to manage the continuous influx of stimuli from the environment. In the retina, receptive fields decrease non-linearly in size from the peripheral retina toward the fovea~\cite{harvey2011relationship}. Consequently, in many mammalian visual systems, the fovea captures a small, highly detailed portion of the scene, while the peripheral retina gathers the remainder at a lower resolution~\cite{Biologyfoveation}. Peripheral vision identifies points of interest and directs the fovea to focus on them sequentially; a process known as \textit{foveation}. This mechanism balances wide-area scanning with a detailed inspection, enhancing efficiency and enabling the recognition of items within the scene~\cite{DigitalFoveation}.

Numerous systems have been proposed to emulate the log-polar mapping and spatially variant sampling distribution characteristic of the visual field~\cite{chessa2016space,GrimesDickeyPishko_2006_EncyclopediaSensors}. This subsampling of visual input offers several advantages, including data reduction~\cite{d2020event}, motion direction detection~\cite{d2020event}, enhanced attention mechanisms, improved object tracking and segmentation, and egomotion segmentation~\cite{BolducLevine_1998_ReviewBiologicallyMotivated}.
The properties of foveated vision have been studied through both software-based simulation and hardware implementations. For instance, as first demonstrated by Geisler \emph{et al.}~\cite{geisler1998real} in 1998 and later confirmed by De Figueiredo \emph{et al.}~\cite{de2023overview}, foveal vision systems are beneficial for reducing computational demand while maintaining effective performance in complex tasks, thereby providing foundational insight into space-variant, non-uniform vision mechanisms. In this regard, Dauce \emph{et al.}~\cite{Dauce_2018_ActiveFoveaBasedVision} have demonstrated that highly accurate active inference can be achieved by applying saccadic movements using only a fraction of the raw visual input by applying a deep neural network as a predictive system. Deep learning models can also achieve both high throughput and high accuracy for saccadic attention through log-polar foveated sampling of the input~\cite{LukanovKonigPipa_2021_BiologicallyInspiredDeep}. An active system with a foveated sensor with log-polar pixel mapping for tracking objects in the visual field~\cite{YamamotoYeshurunLevine_1996_ActiveFoveatedVision} was found to closely replicate the fixation behaviour of humans scanning objects in the visual field. In~\cite{LamTsangMengEtAl_2006_NeuromorphicTranslationalEgomotion}, an FPGA-driven hardware platform for egomotion estimation is implemented on a mobile robotic system, demonstrating that log-polar sampling of the visual input can be harnessed to extract the optical flow and the heading direction of the robot. Comprehensive reviews of foveated systems for robotics can be found in~\cite{BolducLevine_1998_ReviewBiologicallyMotivated,JavierTraverBernardino_2010_ReviewLogpolarImaging}.

Selective attention in computer vision can be approximated by representing the majority of a scene at low resolution while preserving high detail in key regions of interest (ROIs), without implementing any log-polar mapping. To that end, a common approach involves using pairs of sensors to emulate the effects of convergence and disparity estimation~\cite{ChessaMurgiaNardelliEtAl_2014_BioinspiredActiveVision}. For instance, a foveated stereo-matching algorithm~\cite{xu2016evaluation} can significantly improve the efficiency of robotic cloth manipulation tasks, such as grasping and flattening, by achieving comparable accuracy at two to three times the speed of traditional methods. In~\cite{Scassellati_1999_BinocularFoveatedActive}, a binocular foveated system is implemented on a robotic head that can simulate saccadic motion and smooth pursuit, albeit with a significant cumulative pixel error. Additionally, Medeiros \emph{et al.}~\cite{medeiros2020dynamic} introduced a dynamic multifoveated structure that allows robots to track and focus attention on multiple objects simultaneously without redundancy in processing, thereby optimizing real-time image analysis. This technique enhances visual attention capabilities in robotic systems by enabling rapid focus shifts across different regions of interest. Nevertheless, despite the advantages of visual processing with coupled sensors, such as superior object segmentation, considerable effort has been invested in achieving similar results with a single sensor due to the additional complexity necessary for synchronisation and parallel processing when using multiple sensors.

Recent advancements in selective attention systems for neuromorphic vision have highlighted their potential to enhance visual processing efficiency in robotics due to their low power consumption and low latency. 
A pivotal work~\cite{gruel2023stakes} introduces neuromorphic foveation, using event cameras to selectively process visual information, reducing data load while maintaining high accuracy in tasks like semantic segmentation and classification. This approach achieves a superior balance between data quantity and quality compared to traditional high or low-resolution event data.
While foveated vision enhances certain aspects of perception, it introduces an inherent challenge in event processing: distinguishing between independent object motion (\emph{i.e.}, object motion) and camera-induced motion (\emph{i.e.}, egomotion) while executing fixational eye movements and identifying the next salient point for foveation. Disambiguating visual stimuli and perceptually grouping the information to focus the attention requires mechanisms to enhance object motion cues akin to biological vision systems, to ensure reliable object detection and tracking.

Risi \emph{et al.}\cite{9948687} introduced an event-based FPGA implementation of a saliency-driven selective attention model that efficiently processes visual information from event cameras. This implementation outperforms classic bottom-up models, such as those by Itti \emph{et al.}\cite{itti1998model, itti2001computational}, in predicting eye fixations, achieving high performance with minimal computational resources. It captures significant aspects of visual perception by focusing on localised temporal changes and provides microsecond-precise saliency updates without relying on traditional image frames.
Event-based visual attention for robotic applications has already been proposed~\cite{rea2013event}, harnessing its responsiveness and robustness under varying lighting conditions while mimicking human-like attentional processes. 

The turning point in advancing selective attention for object detection was marked by the integration of `Gestalt laws' into bottom-up saliency mechanisms~\cite{russell2014model}, ensuring the perceptual grouping of closed contours as potential objects within the scene.
The frame-based proto-object implementation~\cite{russell2014model} has been further adapted for event-based cameras~\cite{iacono2019proto} for the humanoid robot iCub reducing computational load by eliminating the initial processing stage of edge extraction achieved by utilising the inherent capabilities of the event-based cameras, which interpret the scene with leading and trailing edges, along with polarity, and therefore can be assumed to represent an edge map of the scene. The system ignores clutter and non-proto-object shapes while providing an online saliency map approximately every $\sim$100 ms.
A sparse event-based depth implementation of the model~\cite{ghosh2022event} has shown a more localised response over frame-based implementation~\cite{hu2016proto} ensuring a task-dependent response prioritising proto-objects closer to the observer generating the event-based disparity map and the saliency map approximately every $\sim$170 ms. The final spiking-based implementation of the model~\cite{Dangelo_etal22}, directly deployed on the neuromorphic SpiNNaker platform, achieves a significant reduction in latency (approximately $\sim$16 ms), although it requires an unmanageable number of platforms (6 SpiNNaker boards) to run the full implementation across different orientations and scales of the von Mises filters.

The current challenge is to demonstrate, test, and validate, what is, to the best of the authors' knowledge, the first end-to-end bioinspired attention system leveraging selective attention mechanisms through object motion sensitivity for neuromorphic architectures. The proposed event-based architecture enables an agent to visually explore the scene, using saliency-based proto-object mechanisms to detect and saccade toward potential objects while enhancing relative object motion during fixational eye movements. 
It not only incorporates two fundamental mechanisms occurring in vision during selective attention but also demonstrates the capability of bioinspired software implementation on a robotic PTU.
To avoid introducing an additional layer of complexity, we do not incorporate the spatially non-uniform architecture of the retina in this approach. Instead, we focus on saccadic movements that place the salient points at the center of the visual field.
The entire pipeline operates without requiring any form of training, further demonstrating the power of hierarchical architectures and their robustness across different scenarios. This approach leaves room for the integration of more complex learning algorithms to address advanced tasks in the future.

\section{Methods}

Figure~\ref{fig:pipeline} provides an overview of the entire architecture. The Speck device is mounted on a FLIR pan-tilt robotic unit (VERSION: 3.02). A stream of events is generated by the DVS from the Speck device. These events are directly processed by the spiking Object Motion Sensitivity (sOMS) model, which produces the OMS map as the final output, enhancing object motion. The OMS map is subsequently fed into the SNN Proto-object model to detect proto-objects and identify the most salient point in the scene. The pixel coordinates of the maximum salient point $P(x,y)$ are then input into the SNN Attention Control (sAC) population, which generates the appropriate pan and tilt commands to foveate on the salient point. The system then performs fixational eye movements generating events to detect the next salient point. The simulation is executed on macOS 15.1.1 with an M2 chip, utilizing the Metal Performance Shaders (MPS) device. The system employs the \textit{sinabs} library from SynSense for visual processing and Nengo~\cite{bekolay2014nengo}\footnote{\href{https://www.nengo.ai}{https://www.nengo.ai}} for the control mechanism. To facilitate future deployment on neuromorphic hardware and bridge the gap between simulation and real-world implementation, the Speck device is mounted on the PTU, using only events from the DVS sensors. This setup serves as an intermediate step toward full deployment, enabling us to demonstrate the architecture's capabilities.


\subsection{Spiking Object Motion Sensitivity (sOMS) model}\label{ch:soms}
\begin{figure*}[ht!]
    \centering
    \includegraphics[width=.9\textwidth]{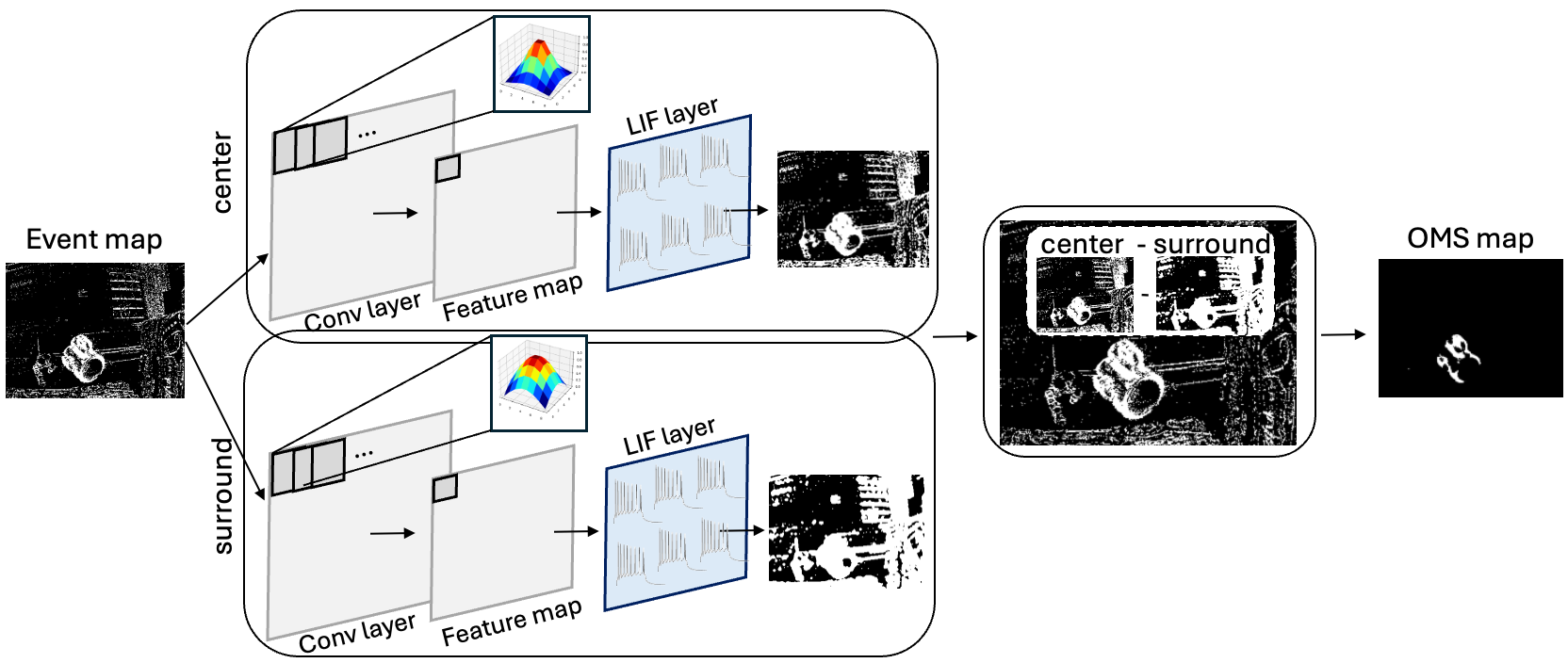}
    \caption{From left to right, an abstract representation of the networks to generate the OMS map. The Event map is processed through the center and surround networks consisting of a convolutional (Conv) layer followed by a LIF layer, producing the intermediate maps used to generate the OMS map, as shown in Equation~\ref{eq:OMSmap}.}
    \label{fig:OMSkernel}
\end{figure*}

The sOMS model receives input from the events generated by the DVS on the Speck device, which has a 128$\times$128 pixel resolution. Central to this implementation are the OMS kernels inspired by the approach outlined in~\cite{clerico2024retina}, which are implemented as two Gaussian distributions, using kernels of 8 pixels (see Figure~\ref{fig:OMSkernel}) and different $\sigma$ for the centre (c) and surround (s) kernels. Unlike the implementation proposed by Clerico \emph{et al.}~\cite{clerico2024retina}, the model simulates the behaviour of OMS cells using the \textit{sinabs} library from SynSense.

The model employs center (c) and surround (s) Gaussian distributions, both centered at \(\mu_{c} = \mu_{s} = (0, 0)\), with the standard deviation \(\sigma_{c} = 1\) and \(\sigma_{s} = 4\) (see Figure~\ref{fig:OMSkernel}). These distributions are implemented through their respective weights through a Leaky Integrate-and-Fire (LIF) Spiking Convolutional Neural Network (sCNN) from the SynSense library, \textit{sinabs}. The network consists of two parallel pathways, center and surround. Each pathway includes a convolutional layer with a Gaussian kernel, followed by a layer of LIF neurons with a membrane time constant \(\tau = 0.02 s\) (see Figure~\ref{fig:OMSkernel}). Motion sensitivity in the model arises from the temporal integration properties of LIF neurons, which enable responses to spatio-temporal activation patterns, facilitating the detection of local contrastive motion.
The events are collected and then fed into the sOMS model, which updates at an empirical rate of \(20 
ms\). This update rate can be reduced to a minimum of \(1 ms\) for real-time processing in the online demonstrator. The collected events, assumed to represent the edge map of the scene (the Event map), are processed through the center and surround kernels. The resulting center map is then subtracted from the surround map (Equation~\ref{eq:OMSmap}).
The final OMS maps for the positive and negative polarities are then obtained similarly to Clerico \emph{et al.}~\cite{clerico2024retina}, as the thresholded version of the initial Event map (Equation~\ref{eq:OMSmap}). 
For simplicity, throughout the remainder of this paper, we refer to their sum, \emph{i.e.}, the combined positive and negative polarity maps, as the OMS map.
However, unlike their approach, we retain the signed difference between the center and surround filters to preserve the center-surround structure, which would be lost if we used the absolute value.

\begin{equation}\label{eq:OMSmap}
\centering
\begin{aligned}
    OMS_{map} &= \text{center}(Events_{map}) - \text{surround}(Events_{map}) \\
    OMS_{map} &= Events_{map} \left[ 1 - \text{norm}(OMS_{map}) \geq \alpha \right]
\end{aligned}
\end{equation}

Where $\alpha$ acts as a threshold that depends on the visual input.
As shown in Figure~\ref{fig:pipeline}, the output of the sOMS model generates an OMS map that emphasises sensitivity to object motion, or, in other words, emphasises local contrast in motion.
The OMS map is then input into the SNN Proto-object model to detect proto-objects within the scene, ensuring the identification of a salient region that may contain an object.

\subsubsection{Characterisation of the sOMS model}\label{ch:charOMS}

\begin{figure*}[ht!]
    \centering
        
        
    \includegraphics[width=1\textwidth]{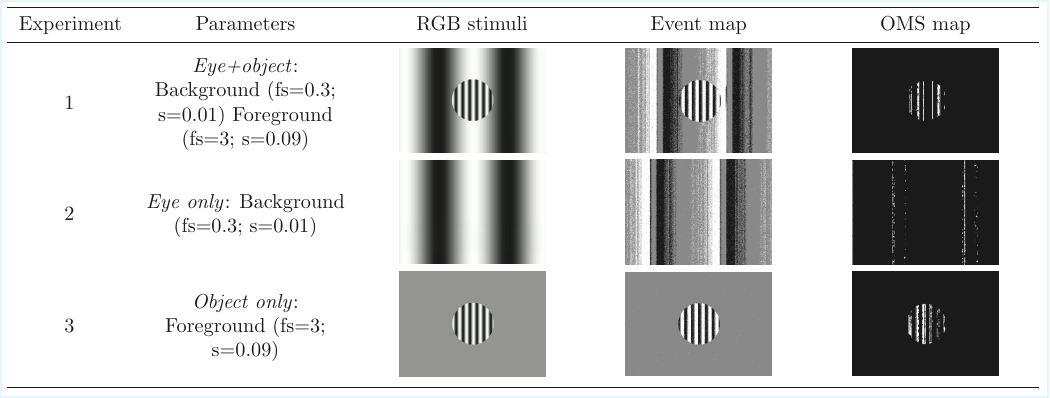}
\caption{From left to right: Experiment, Parameters, RGB stimuli, Event map and OMS map in three different situations: \textit{Eye+object}, \textit{Eye only} and \textit{Object only}. Where ${fs}$ is the spatial frequency in cycles per visual degree (deg) and ${s}$ is the speed in cycles per frame of the moving grating. The figure shows the sOMS model enhancing object motion, where white and black represent positive and negative polarities on the Event map, and white indicates the spikes of the sOMS model on the sOMS map.}
\label{fig:OMSchar}
\end{figure*}

The characterisation of the sOMS has been inspired by the recordings of real cells conducted by {\"O}lveczky \emph{et al.}~\cite{olveczky2003segregation}, where the activity of cells was measured under various conditions: \textit{Eye+object}, \textit{Eye only} and \textit{Object only}. The stimuli consist of moving gratings, with both the background and central region in motion. In the case of \textit{Eye+object}, both the background and the central region move at different speeds; in \textit{Eye only}, only the background is moving; and in \textit{Object only}, only the central part is moving (see Figure~\ref{fig:OMSchar}).


The stimuli have been generated using the PsychoPy library~\cite{peirce2007psychopy}. 
The gratings for both the background and the central object move with a speed $s$ and a spatial frequency $fs$. 
The stimuli presented to the model are RGB video clips with a duration of $4s$ at $60$ fps. The stimuli are then converted into events using the ICNS Event-Based Camera Simulator (IEBCS)\footnote{\href{https://github.com/neuromorphicsystems/IEBCS}{https://github.com/neuromorphicsystems/IEBCS}} library using the following standard parameters of the simulator: ${dt} = 10000$ $\mu$s, ${lat} = 100$ $\mu$s, ${jit} = 10$ $\mu$s, ${ref} = 100$ $\mu$s, ${tau} = 300$ $\mu$s, ${th} = 0.1$ and $th_{noise} = 0.01$.
The IEBCS library generates realistic event-based camera data by processing video frames to detect changes in pixel brightness over time, then producing asynchronous event streams that include precise timing, sensor-specific noise, and latency effects to closely mimic real neuromorphic vision sensors.
Where ${dt}$ is the delta time, ${lat}$ is the latency, ${jit}$ the jitter, ${ref}$ the refractory period, ${tau}$ the time constant, ${th}$ the positive/negative events generation threshold, and  $th_{noise}$ is the added noise sampled from two distributions acquired using a real sensor under 161 lux.
Figure~\ref{fig:OMSchar} illustrates an example of the RGB stimuli, the converted event frame, and the OMS maps in the three different cases: \textit{Eye+Object}, \textit{Eye Only} and \textit{Object Only}. 
The sOMS model encodes object motion sensitivity~\cite{clerico2024retina,olveczky2003segregation}: it produces a robust response when the central object moves against background motion in the \textit{Eye+Object} scenario and, as expected, in the \textit{Object Only} scenario. 


\begin{figure}[htbp]
    \centering
    \includegraphics[width=1\textwidth]{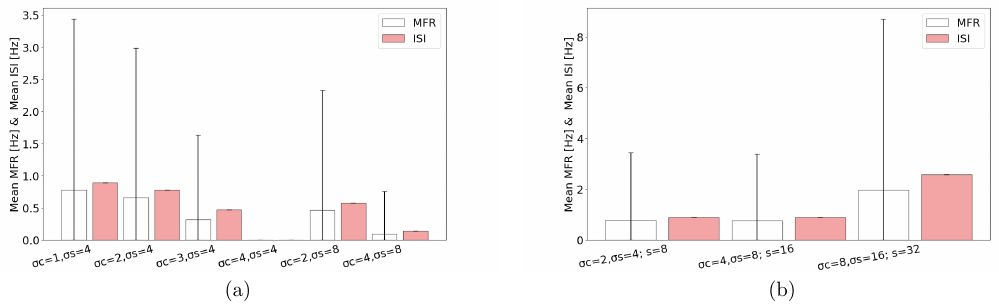}
    \caption{Average mean firing rate (MFR) and mean inter-spike interval (ISI), along with their respective standard deviations, for the \textit{Eye+Object} case under two scenarios: (a) varying center ($\sigma_{c}$) and surround ($\sigma_{s}$) values of the kernels (pixels) [$\sigma_{c}=1,\sigma_{s}=4$; $\sigma_{c}=2,\sigma_{s}=4$; $\sigma_{c}=3,\sigma_{s}=4$; $\sigma_{c}=4,\sigma_s=4$; $\sigma_{c}=2,\sigma_{s}=8$; $\sigma_{c}=4,\sigma_{s}=8$] (Figure~\ref{fig:OMSkernel}), and (b) varying both $\sigma$ values and kernel sizes [$\sigma_{c}=1,\sigma_{s}=4$, s=8; 
    $\sigma_{c}=4,\sigma_{s}=8$, s=16;
    $\sigma_{c}=8,\sigma_{s}=16$, s=32].}
    \label{fig:sigmakernels}
\end{figure}


The Gaussian kernel parameters and sizes were analysed based on their responses in the \textit{Eye+Object} scenario. Figure~\ref{fig:sigmakernels}a) illustrates how decreasing the ratio between $\sigma_{c}$ and $\sigma_{s}$ proportionally reduces both the mean firing rate (MFR) and the mean inter-spike interval (ISI) of the system, due to the diminished inhibitory effect of the surround. These metrics were selected as they provide interpretable and reliable descriptors of spiking activity under varying stimulus conditions.
The final two edge cases, in which $\sigma_{s}$ equals the kernel size, show a slight reduction in both metrics, leading to an overall decrease in spiking activity as a result of stronger surround suppression.
Larger kernel sizes, as shown in Figure~\ref{fig:sigmakernels} b), confirm the same response trend, with increased activity observed for the largest size. This effect is attributed to the smoothing influence of the Gaussian kernels on the image, resulting in a larger portion of the visual field surpassing the threshold ($\alpha$). An increased MFR would lead to heightened system activity, resulting in unnecessary power consumption, that in neurons chips scales with the number of spikes, as the response is already functional for $\sigma_{c}=1$, $\sigma_{s}=4$, and a kernel size, $s_{k}$, of 8 pixels. This analysis, therefore, validates the parameters proposed in~\cite{clerico2024retina}.

\begin{figure*}
    \centering
    \includegraphics[width=1\textwidth]{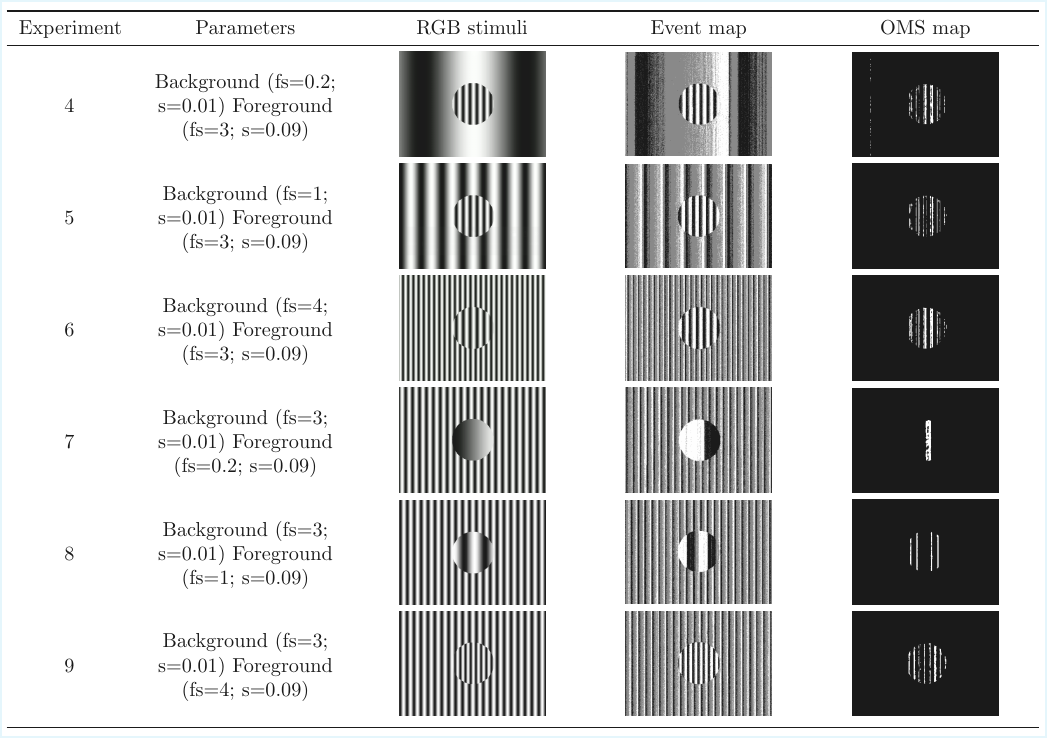}
\caption{From left to right: Experiment, Parameters, RGB stimuli, Event map and OMS map for the \textit{Eye+object} case with different spatial frequencies of the background and the foreground. Where ${fs}$ is the spatial frequency and ${s}$ is the speed in cycles per frame of the moving grating. The colors shown on the maps are the same as those in Figure~\ref{fig:OMSchar}.}
\label{fig:OMScharsize}
\end{figure*}

To demonstrate the scale invariance of the model, the characterisation also includes different spatial frequencies of the stimuli in the \textit{Eye+Object} scenario. Figure~\ref{fig:OMScharsize} shows an example of the RGB stimuli, the converted event frame, and the OMS maps with different values of spatial frequency, ${fs}$, and the speeds of a cycle on each iteration of the loop, ${s}$, for the object (Foreground) and the Background. As shown by the OMS maps in the qualitative assessment in Figure~\ref{fig:OMScharsize}, the response of the sOMS model effectively enhances object motion in all cases.


\begin{figure*}[ht!]
    \centering
    \includegraphics[width=0.8\textwidth]{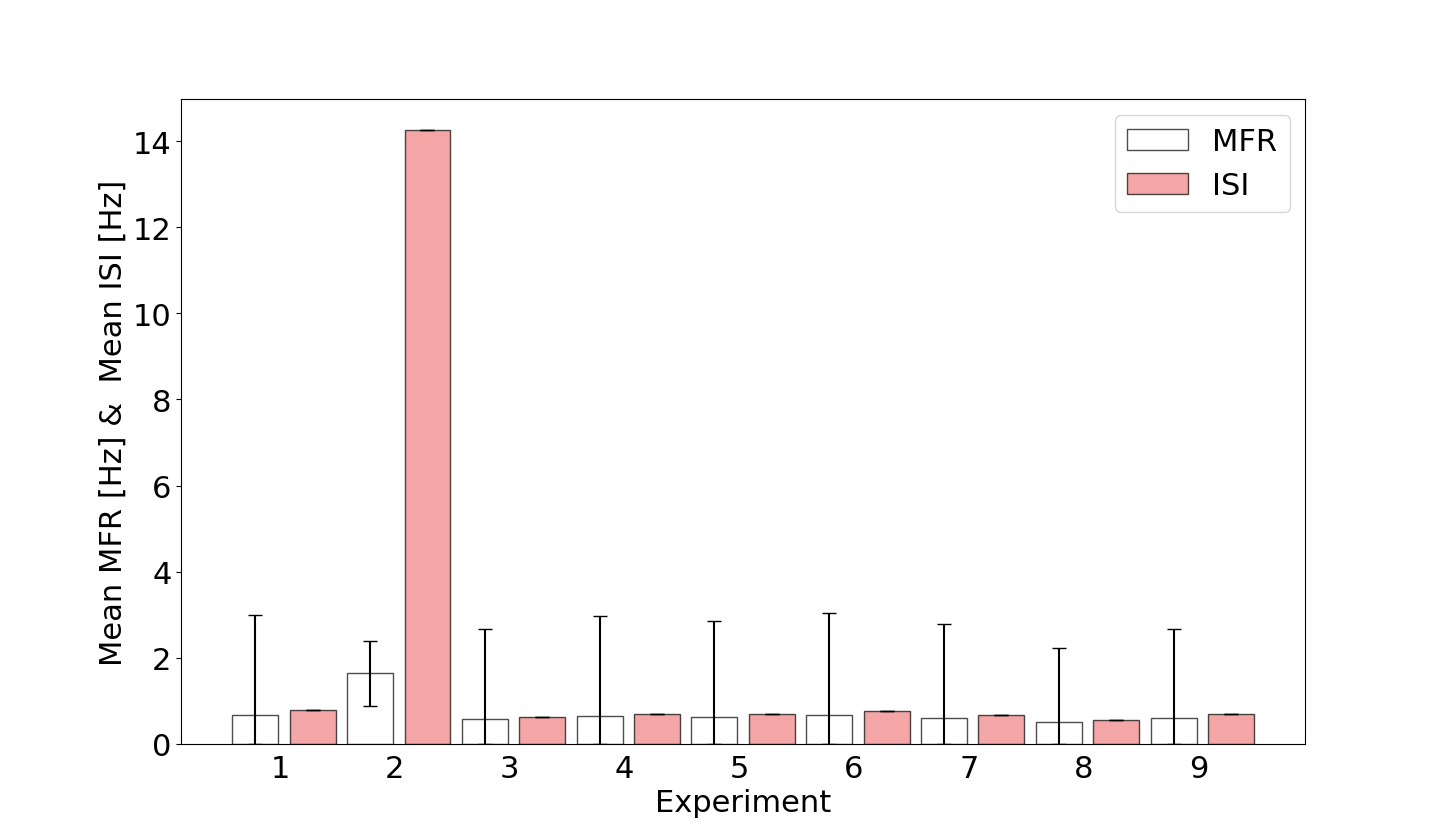}
    \caption{Average MFR and ISI, along with their respective standard deviations, are reported for all experiments shown in Figure~\ref{fig:OMSchar} and Figure~\ref{fig:OMScharsize}. Experiments 1, 2, and 3 correspond to \textit{Eye+Object}, \textit{Eye only}, and \textit{Object only}, respectively, while experiments 4 through 9 represent \textit{Eye+Object} scenarios with varying spatial frequencies (${fs}$) and speeds (${s}$) in cycles per iteration, as depicted in Figure~\ref{fig:OMScharsize}.}
    \label{fig:OMSMFRISI}
\end{figure*}

Further investigations analyze the average MFR and mean ISI of the population activity (see Figure~\ref{fig:OMSMFRISI}) across all experiments presented in Figure~\ref{fig:OMSchar} and Figure~\ref{fig:OMScharsize}, as these metrics provide complementary insights into the dynamics of neural activity. The average MFR captures the overall firing rate, offering a measure of activity intensity, while the mean ISI reflects the temporal structure of spiking, highlighting patterns in neural coordination and timing. The responses confirm the periodicity of the moving gratings, with similar MFR and mean ISI observed across all scenarios, except for Experiment 2 (\textit{Eye only}). 
In this case, the OMS model remains effective in suppressing most background activity, even in the absence of object motion; however, the periodicity of the response is lost. This results in a higher mean MFR compared to the other conditions and an elevated mean ISI, indicating a stronger spatial response from the neurons throughout the simulation period, which was not evident in the single example shown in Figure~\ref{fig:OMSchar}.
The standard deviation of the mean MFR reflects that only a subset of neurons, corresponding to a specific portion of the visual field, are actively responsive across all cases. In contrast, the standard deviation of the mean ISI becomes extremely large (approaching infinity), as most neurons emit only a single spike during the simulation, resulting in undefined or extremely high interspike intervals. 


\begin{figure*}[ht!]
    \centering
    \includegraphics[width=1\textwidth]{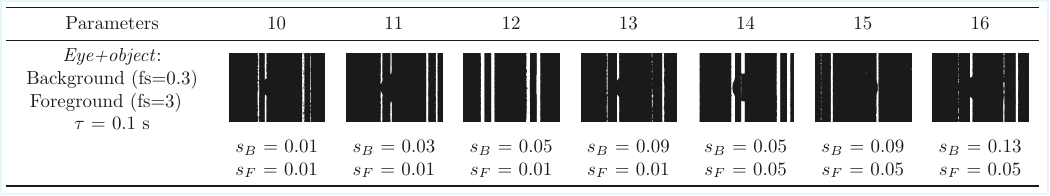}
\caption{Example of OMS map for different speeds (where $s_{F} \leq s_{B}$) for the background ($s_{B}$) and foreground ($s_{F}$) in the case of the \textit{Eye+object} scenario for the Background ($fs$=0.3) and the Foreground ($fs$=3).}
\label{fig:speedchar}
\end{figure*}

In all the aforementioned cases, the speed of the foreground is greater than that of the background. To better understand the role of speed in foreground and background motion, additional investigations in the \textit{Eye+object} scenario were conducted to examine the effects of varying grating stimuli speeds (see Figure~\ref{fig:speedchar}). In this scenario, the background motion is faster than the foreground, indicating faster egomotion-induced events and slower foreground motion.


In all cases examined within Experiments 10–16 (Figure~\ref{fig:speedchar}), the sOMS model appears unable to enhance object motion effectively (same applies for $\tau$ = 0.01), resulting in poor performance in object motion segmentation when egomotion-induced events are generated from a background moving faster than the foreground.
The failure of motion detection can be attributed to two critical mechanisms.
First, spatiotemporal integration between center and surround regions becomes ineffective when object-induced events are temporally sparse or significantly lower than background-induced events. Under such conditions, the membrane potential fails to reach the threshold required for motion detection or the dominance of background activity suppresses the differential signal needed to distinguish motion.

Second, from an active vision perspective, the motion dynamics of the visual sensor critically impact detection reliability. 
The challenge of motion detection under rapid camera movement is a well-documented issue in optic flow estimation~\cite{pauwels2006optic}.
When camera movement exceeds a specific velocity threshold, local motion contrast becomes unresolvable, directly compromising the fundamental assumptions of optical flow estimation.

These constraints underscore the intricate relationship between spatio-temporal processing and motion. This provides insights into the model's ability to accurately segment motion in scenarios where the background moves more slowly than the foreground. Alternatively, it evaluates the model's performance from an active vision perspective, where the vision sensor moves at a slower speed than the dynamic scene, highlighting the need for small adjustment movements, such as fixational eye movements, to function properly.

\subsection{SNN Proto-Object model}\label{subsec:snn-proto-object}

This implementation builds upon the event-driven, saliency-based proto-object model previously developed for GPU~\cite{iacono2019proto} and SpiNNaker~\cite{Dangelo_etal22}. The primary difference from the earlier implementations is the use of the LIF sCNN \textit{sinabs} layer, with weights represented by the von Mises kernel distribution (Equation~\ref{eq:VMeq}).

\begin{equation} \label{eq:VMeq}
    VM_{\theta} (x,y) = \frac{\exp(\rho \cdot R_0 \cdot \cos(\tan^{-1}(-y, x) - \theta)}{I_0(\sqrt{x^2 + y^2 - R_0})}
\end{equation}

Where $x$ and $y$ are the kernel coordinates with the origin at the center of the filter, $\theta$ represents its orientation, $I_0$ is the modified Bessel function of the first kind, and $R_0$ is the radius of the filter, which represents the distance of the VM filter from the center. This radius will later define the radius of the opposite orientation of the von Mises filters, grouping information to detect the proto-object and defining the range of detectable sizes. The function $\tan^{-1}$ takes two arguments and returns values in radians in the range $(-\pi, \pi)$.

As described in Equation~\ref{eq:BO}, the model computes the Border Ownership layer using the OMS map from the sOMS layer (a binary matrix $V$) as input. It processes the positive ($V_+$) and negative ($V_-$) events with four kernel orientations and their corresponding opposite orientations: $\theta = [0\degree, 45\degree, 90\degree, 135\degree]$.

\begin{equation} \label{eq:BO}
     \begin{aligned}
        B1_\theta = V &\odot (V_+ * VM_\theta - w V_- * VM_{\theta + \pi} \\
        &+ V_- * VM_\theta - w V_+ * VM_{\theta + \pi} ) \\
        B2_\theta = V &\odot (V_+ * VM_{\theta + \pi} - w V_- * VM_{\theta} \\
        &+ V_- * VM_{\theta + \pi} - w V_+ * VM_{\theta} )
     \end{aligned}
\end{equation}

Here, 
$*$ denotes convolution and $\odot$ indicates element-wise multiplication. The factor~$w$ modulates the inhibition between competing polarities and orientations. The higher its value, the more difficult it becomes for one orientation to dominate its opposite, effectively reducing ambiguities.

\begin{equation}
    \begin{minipage}{0.45\textwidth}
        \centering
        $G_1 = \sum_{\theta} \sum_{s} B1_{\theta, s} \cdot VM_{\theta, s}$\\
        $G_2 = \sum_{\theta} \sum_{s} B2_{\theta, s} \cdot VM_{\theta + \pi, s}$
    \end{minipage}
    \hfill
    \begin{minipage}{0.45\textwidth}
        \centering
        $G_1^* = \sum_{\theta} \sum_{s} B1_{\theta, s} \cdot VM_{\theta + \pi, s}$\\
        $G_2^* = \sum_{\theta} \sum_{s} B2_{\theta, s} \cdot VM_{\theta, s}$
    \end{minipage}
    \label{eq:G}
\end{equation}

These orientations and scales ($s = 3$) are pooled (Equation~\ref{eq:G}) to produce the final grouping response, which generates the saliency map of the scene and identifies the $(x, y)$ coordinates of the most salient point (Equation~\ref{eq:pxy} and Figure~\ref{fig:pipeline}).

\begin{equation}\label{eq:pxy}
    P(x, y) = \arg\max_{(x, y)} \left[ (G_1 - G_1^*) + (G_2 - G_2^*) \right]
\end{equation}

These coordinates, $P(x, y)$, are subsequently passed to the Spiking Attention Control model to generate a saccade toward the salient point.

\subsubsection{Characterisation of the SNN Proto-Object model}


\begin{figure*}[ht!]
    \centering
    \includegraphics[width=1\textwidth]{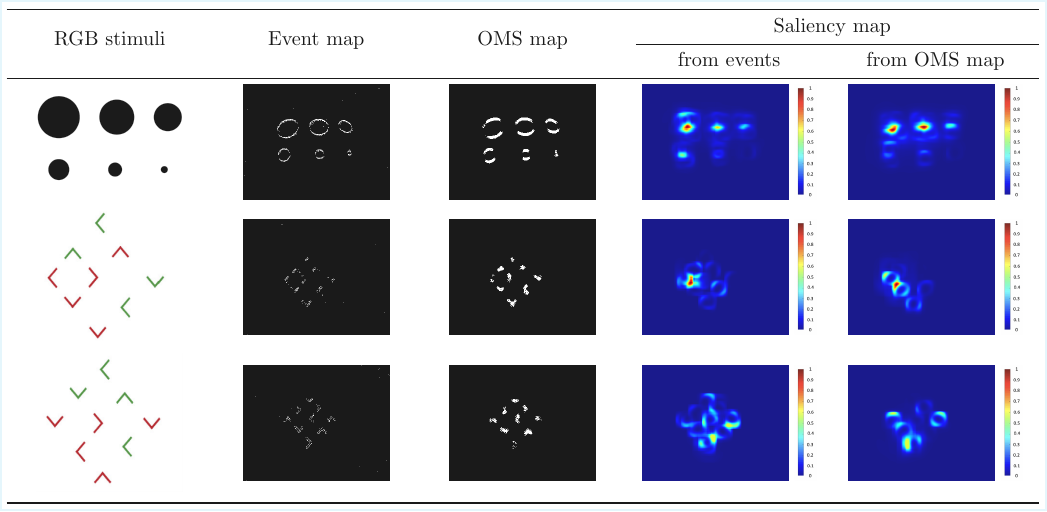}
\caption{From left to right: RGB stimuli in the case of \textit{calibration circles}, \textit{proto-object} patterns, and \textit{non-proto-object} patterns from the SalMapIROS dataset~\cite{iacono2019proto}, Event map, OMS map, and Saliency map (the color legend follows the jet map, with blue indicating minimum values and red representing maximum values) with events and OMS map as input, respectively.}
\label{fig:attchar}
\end{figure*}


To validate the system's functionality, we replicated the characterisation described in~\cite{iacono2019proto,ghosh2022event,Dangelo_etal22}. Events from the SalMapIROS dataset~\cite{iacono2019proto} were utilised to qualitatively evaluate the system's performance across three primary scenarios: \textit{calibration circles} (ideal stimuli), \textit{proto-object} patterns, and \textit{non-proto-object} patterns (see Figure~\ref{fig:attchar}).
The \textit{proto-object} pattern derives from the characterisation of the original RGB implementation~\cite{russell2014model} depicting corners and corners facing each other, which conceptually resembles a proto-object.

In~\cite{iacono2019proto} the events were collected using ATIS cameras mounted on the humanoid robot iCub~\cite{bartolozzi2011embedded} by generating small stereotypical circular eye movements, emulating fixational eye movements observed in many vertebrate species~\cite{martinez2008fixational}. During the experiments, the robot observed patterns placed on a desk (72 cm high), while the camera setup was positioned at a total height of approximately 124 cm (104 cm for the robot's base height plus ~20 cm for support).
This range of sizes was selected based on the typical objects the robot can interact with, considering the constraints of its workspace and grasping capabilities~\cite{vezzani2017grasping, jamone2016benchmarking}.
The characterisation analyses the responses feeding the SNN Proto-object model with raw events and the OMS map from the sOMS model. 

In the \textit{calibration circles} scenario, the system, with both inputs, detected almost four out of six circles, with radii ranging from 10 to 60~mm, corresponding to sizes between 25 and 50 pixels on the image plane. In the \textit{proto-object} pattern, the model exhibited a strong response to corners facing each other representing proto-objects~\cite{russell2014model} while displaying a weaker response to corners in the \textit{non-proto-object} pattern.
Furthermore, the OMS map, due to Gaussian filtering, produces a thicker and clearer edge map, making it an ideal input candidate for the SNN Proto-object model. The parameters obtained from the characterisation are summarised in Table~\ref{table:params}.

\subsection{Spiking Attention Control}

The Spiking Attention Control (sAC) obtains $P(x, y)$ from the SNN Proto-object model and generates the saccadic movements (pan-tilt) through a simple attention control model implemented using spiking neurons. The objective of the control system is to minimise the distance between the centroid of the most salient object and the center of the sensor's field of view. This is achieved by moving the center of the DVS integrated into the Speck device mounted on the FLIR PTU to \textit{foveate} towards the salient point (see Figure~\ref{fig:pipeline}). 
To describe this feedback control system, we use the Neural Engineering Framework (NEF)~\cite{eliasmith2003neural}. Since the camera is mounted on a PTU, we can treat the axis-aligned errors and controls separately. A proportional controller is used, defined as:

\begin{equation}
    \begin{bmatrix}
        u_\mathrm{pan}\\
        u_\mathrm{tilt}
    \end{bmatrix} = \begin{bmatrix}
        K_\mathrm{pan} & 0 \\
        0 & K_\mathrm{tilt}
    \end{bmatrix}\begin{bmatrix}
        x_\mathrm{obj} - x_\mathrm{center}\\
        y_\mathrm{obj} - y_\mathrm{center}
    \end{bmatrix}.
\end{equation}

The neural controller was implemented using spiking LIF neurons within the Nengo software package~\cite{bekolay2014nengo} and was instantiated by solving a least-squares optimization problem to determine the decoder weights, $\mathbf{w}$, of a single-hidden-layer neural network, assuming fixed, randomly generated input weights, $\mathbf{e}$. For the pan controller, the following equation is solved:

\begin{equation}
    \mathbf{w}_\mathrm{pan} = \underset{\mathbf{w} \in \mathbb{R}^{d}}{\arg\min} \left\|\frac{K_\mathrm{pan}}{\text{degrees\_per\_pos} } \varepsilon_{x} - \mathbf{w} \cdot \left(h(t)\ast \mathbf{a}(\langle \mathbf{e}_\mathrm{pan},\varepsilon_{x}\rangle)\right)\right\|^{2}
    \label{eq:wpan-training}
\end{equation}

where $h(t)$ is a synapse model (a low-pass filter), $a(\cdot)$ is the activity for the $d$ leaky integrate and fire neurons (bias term omitted for compactness) in the population representing the pan error, $\mathbf{e}_\mathrm{pan}$ are the randomly generated encoding weights for the pan controller, and $\varepsilon_{x} = x_\mathrm{obj}-x_\mathrm{center}$, is the control error given the current saliency input. Here we make explicit the term $\text{degrees\_per\_pos}$, which is the angular resolution. The same technique is used to solve for the tilt control weights, $\mathbf{w}_\mathrm{tilt}$. We exploit Kirchoff's current laws to compute the difference between the actual and desired location of the saliency maximum directly in the current injected into the neurons. For a fixed target location, this difference could also be encoded in the biases of the neurons that make up the controller.

To avoid interference between the two control dimensions, a separate neural population is maintained for both the pan and tilt commands. Proportional control gains were hand-tuned to produce the desired performance. 
The final values of $cmd_{\text{pan}}$ and $cmd_{\text{tilt}}$ are obtained from the $u_{\text{pan}}$ and $u_{\text{tilt}}$ output commands of the sAC control as follows:

\begin{equation}\label{eq:upan}
    \text{cmd}_{\text{pan}} = \left\lfloor u_{\text{pan}} \right\rfloor,
\end{equation}

where $\lfloor \cdot \rfloor$ represents the floor operator. The same relation applies for $u_{tilt}$. While this work employs a simple proportional controller to direct eye movements, these neural control techniques can be extended to more complex feedback controllers, including adaptive components (\emph{e.g.},~\cite{dewolf2016spiking,damberger2025adaptiveSSP}), although this is beyond the scope of the current work.

\subsubsection{Characterisation of the Spiking Attention Control}

To evaluate the accuracy of the sAC model in foveating towards a desired point, 10 trials were conducted, and the displacement in \(x\) and \(y\) pixel coordinates from the sAC pixels commands, $cmd_\text{pan}\cdot\text{degrees\_per\_pos}$ and $cmd_\text{tilt}\cdot\text{degrees\_per\_pos}$, to the salient point $P(x,y)$, were analysed. The results showed a mean displacement of \(11.76\) pixels (standard deviation: \(2.58\)) in the vertical (\(y\)) direction and \(4.64\) pixels (standard deviation: \(1.49\)) in the horizontal (\(x\)) direction, indicating greater accuracy in the horizontal axis. 

The greater displacement variability observed in the vertical $y$ direction can be attributed to a combination of mechanical imprecision and gravitational effects. The tilt mechanism of the PTU is more susceptible to play and backlash compared to the pan mechanism, leading to reduced accuracy in vertical movements. Furthermore, the camera is mounted with an offset with respect to the PTU's horizontal hinge, which produces two effects that exacerbate the error in the tilt direction, further compromising vertical accuracy. The first effect is mechanical, whereby the offset physically introduces torque effects on the hinge. The second effect is caused by the fact that the center of the camera sensor itself moves in the vertical direction as the unit tilts. These factors collectively explain the observed discrepancy in displacement precision between the vertical and horizontal directions. We anticipate that integrating an adaptive controller in future work to compensate for unmodelled constraints on the actuator.

\subsection{Real-world demonstrator: Bioinspired attention system}

The real-world implementation has been developed (Figure~\ref{fig:pipeline}) to streamline the data flow and maximise computational efficiency. 
Some optimisations are crucial for handling high-frequency event data, commonly encountered in real-time applications such as robotic vision. Parallel processing for pan-tilt movement management, noise reduction, and the minimisation of redundant computations enables responsive and accurate real-time performance, even with large volumes of event data.
Below, we outline the key aspects of our implementation, demonstrating the operation of the proposed event-based visual attention pipeline\footnote{The code can be found at \href{https://github.com/GiuliaDAngelo/Speckegomotion/tree/demo}{https://github.com/GiuliaDAngelo/Speckegomotion/tree/demo}}.

The pipeline consists of several initialisation steps followed by a loop that processes events arriving from the event source. The \textit{object motion sensitivity} module, the \textit{proto-object detection} module, the \textit{visual attention} module with the Nengo controller and the PTU are initialised with the parameters outlined in Table~\ref{table:params}.
Additionally, the number of neurons ($n$) in the system's convolutional layers is directly related to the number of units in the output feature map, which is determined by the input size, kernel size, and stride.

To ensure compatibility with the Speck device for future deployment and optimise the system's performance for real-time demonstration on the MPS, we have implemented a streamlined version of the SNN Proto-object model. This version eliminates the polarity split by convolving with different orientations and pooling the information to produce the saliency map. Although this simplified attention model does not yet operate on neuromorphic hardware, it effectively demonstrates the core functionality and is well-suited for deployment on Speck. This approach allows us to showcase the system's potential in real-world scenarios.

\begin{table}
	\centering
\begin{tabular}{ccc}
	sOMS & SNN Proto-object & sAC\\
	\hline
        $\tau$ = 0.02 s & $\tau$ = 0.1 s & pan population = 50 neurons\\
        $s_{k}$ = 8 pxs & $R_0$ = 8 pxs & tilt population = 50 neurons \\
	$\sigma_{c}$ = 1, $\sigma_{s}$  = 4  & $\rho$ = 0.2 & $(x,y)_{center}$ = 64\\
        $\alpha$ = 0.80 & $\omega$ = 3 & $k_{pan}$=$k_{tilt}$=1\\
	  $n$ = 29282 & Pyramid levels = 3  &\\
	   & Orientations: [$0\degree$, $45\degree$, $90\degree$, $135\degree$] \\ 
        & $n$ = 113512 &\\  
\end{tabular}
\caption{\label{table:params}Values of the parameters used in the sOMS, SNN Proto-object and sAC models.}
\end{table}

The implementation focuses on demonstrating an end-to-end bioinspired attention system that uses event-based data and pan-tilt movements to foveate the most salient point in the visual field, performing fixational eye movements to detect the next point, as shown in Figure~\ref{fig:pipeline}.
To achieve this, the pipeline implements a loop that accumulates events over a short period of time (2 ms) and feeds them into the \textit{object motion sensitivity} module, which outputs and highlights regions of non-coherent motion.

A separate thread is used to handle the \textit{proto-object detection} module, the \textit{visual attention} module, and the \textit{fixational eye movements} module, ensuring that the event source is not blocked.  
The \textit{proto-object detection} module generates a saliency map, which may contain one or more potential salient objects. From this map, a single maximum salient point, \(P(x, y)\), is identified and passed to the \textit{visual attention} module.
\(P(x, y)\) in pixel coordinates are then passed to the Nengo controller, which performs a regression over the coordinates to compute the PTU control command. The PTU command is in the form of pixel offsets from the center. 

The pan and tilt ranges are defined based on the DVS angular resolution, field of view, and physical constraints of the PTU. The angular resolution is determined by the conversion factor \(\text{degrees\_per\_pos} = 0.02572^\circ\), which corresponds to one PTU position of \(92.5714\) arcseconds (\(92.5714 \, \text{arcsec} \times 0.0002778^\circ/\text{arcsec} = 0.02572^\circ\)). The field of view is calculated using the focal length of the DVS sensor \(f = 1.7 \, \text{mm}\) and the horizontal sensor size \(s_{\text{hor}} = 5.12 \, \text{mm}\). The horizontal field of view \(\text{FOV}_{\text{hor}}\) is derived as:
\[
\text{FOV}_{\text{hor}} = 2 \cdot \arctan\left(\frac{s_{\text{hor}}}{2 \cdot f}\right)
\]
The pan range is then computed as half the ratio of the field of view to the angular resolution to allow movements from the center to the right and from the centre to the left:
\[
\text{pan}_{\text{limit}} = \left\lfloor \frac{\text{FOV}_{\text{hor}}}{\text{degrees\_per\_pos}} \right\rfloor // 2.
\]
The resulting range is used to define the pan and tilt limits for the vision sensor:
\[
\text{pan}_{\text{range}} = \left[-\text{pt}_{\text{limit}}, \text{pt}_{\text{limit}}\right]
\]
The same applies for the $\text{tilt}_{range}$ since the resolution of the DVS is 128$\times$128. 
Finally, the pan and tilt ranges are clipped to ensure they remain within the physical limits of the PTU.

Once the PTU has oriented the system's fovea toward the salient point, it initiates a sequence of fixational eye movements, mirroring those observed in biological vision systems~\cite{ko2010microsaccades}. These fixational eye movements are modelled using a random walk mechanism. Specifically, for each movement \(i\) in a sequence of \(N\) movements, both the pan and tilt angles are updated by adding a small random perturbation \(r_i\), drawn from a uniform distribution:

\[
\text{for } i \in \{1, 2, \dots, N\}:
\]
\[
\text{pan\_angle} \leftarrow \text{pan\_angle} + r_{i}, \quad
\text{tilt\_angle} \leftarrow \text{tilt\_angle} + r_{i}.
\]

Here, \(N=8\) and the perturbation range is \([-60, 60]\) degrees. This choice of parameters was empirically determined to effectively generate events from the static scene while providing sufficient time for spatiotemporal integration of the information.
These updated angles are sent as commands to the PTU, simulating the small, involuntary movements characteristic of fixational eye movements and necessary to process the new information of the visual field and detect the next salient point.

\begin{algorithm}
\DontPrintSemicolon

\caption{Algorithm for the Bioinspired visual attention system}\label{alg:OMS-visual-attention}
\KwData{Events}
\KwResult{Command for PTU - pan and tilt angles}
Initialise $OMS\_module$, $attn\_module$ and $Nengo\_controller$\;
\Fn(){PTU\_ctrl\_function(vSliceOMS)}{
    \Comment{Runs in a separate thread or process to prevent blocking the event source}

    $saliency\_map \gets attn\_module(vSliceOMS)$\;
    $max\_saliency\_coords \gets argmax(saliency\_map)$\;
    $cmd\_pan, cmd\_tilt \gets Nengo\_controller(max\_saliency\_coords)$\;
    assert that $cmd\_pan \in pan\_range, cmd\_tilt \in tilt\_range$\;
    \Comment{The following pan and tilt offsets for the PTU are obtained as per Eq.\ref{eq:upan}}
    $u\_pan, u\_tilt \gets cmd\_pan$\;
    issue \textit{blocking} command to PTU unit to move by $u\_pan$, $u\_tilt$\;
    issue \textit{non-blocking} command to PTU to perform small random movements (\textit{fixational eye movements})\;
}

\While{ev\_src produces events} {
    \Comment{Create event frames from the source (an event camera)}
    $event\_frame \gets accumulate\_events(ev\_source)$\;
    \Comment{Create positive and negative OMS maps}
    $OMS \gets oms\_module(window)$\;
    $vSliceOMS \gets stack(OMS)$\;
    PTU\_ctrl\_function(vSliceOMS)\;
}

\end{algorithm}


\section{Experiments \& Results}

The experiments, both qualitative and quantitative, evaluate the system as a whole, quantifying the advantages of utilising the sOMS model for object motion segmentation and enhancing multi-object detection through the SNN Proto-object model.
The sOMS was initially evaluated to quantify the proportion of events suppressed that are unlikely to correspond to moving objects. To assess the model's accuracy in effectively segmenting multiple objects, it was benchmarked on the EVIMO~\cite{mitrokhin2019ev} dataset, measuring mean IoU and mean SSIM percentages.
To further evaluate the contribution of the SNN Proto-object model, we qualitatively compared results with and without its inclusion and quantitatively assessed its accuracy in multi-object detection.
To further leverage the inherent advantages of event-driven cameras, the sOMS model and the SNN Proto-object model have been qualitatively and quantitatively evaluated on the LLE-VOS~\cite{li2024event} dataset under challenging indoor and outdoor lighting conditions,  where we calculated the accuracy of the SNN Proto-object model in object detection.
Finally, the entire closed-loop bioinspired attention system has been demonstrated, with the Speck device mounted on the PTU successfully performing saccades toward the salient point of interest and generating fixational eye movements to detect the next salient point (see the accompanying video\footnote{\href{https://youtu.be/dcAJlDgVR0o}{https://youtu.be/dcAJlDgVR0o}}). 

\begin{figure*}[ht!]
    \centering
    \includegraphics[width=1\textwidth]{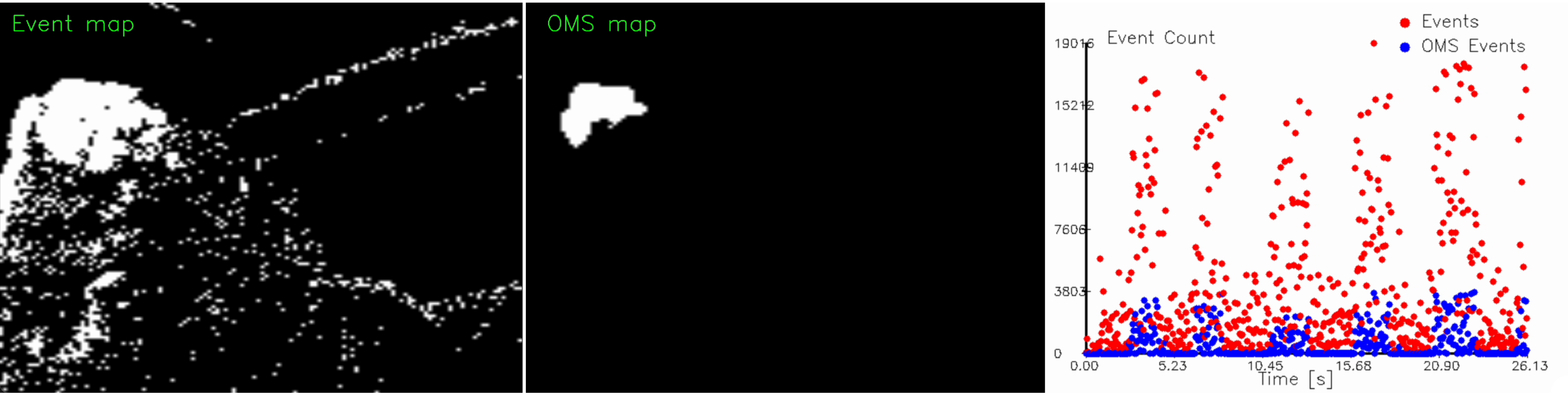}
    \caption{An example from one of the 10 trials illustrating the event counts before and after applying the sOMS model to events received from the DVS sensor (128$\times$128 pixels). The events count from the DVS is represented in red and the outcoming events from the sOMS model in blue. The y-axis shows the event count, while the x-axis represents time in seconds.}
    \label{fig:ev_supp}
\end{figure*}

\subsection{sOMS model: Event suppression}

As described in Section~\ref{ch:charOMS}, the sOMS model enhances motion contrast, allowing objects with relative motion to the background to emerge more distinctly. Simultaneously, it suppresses a substantial number of events unrelated to object motion, such as those caused by static background elements or sensor noise. Unlike generic event-reduction techniques that indiscriminately discard data, the sOMS model reduces events not by simply lowering the count (event-based cameras can potentially generate tens of millions of events per second) but by achieving targeted suppression based on motion relevance. This preserves salient object features while improving the efficiency of downstream processing. As a result, the computational load is significantly reduced without compromising task-relevant information, highlighting a key advantage of the proposed model. 
To quantitatively validate this observation, the number of events was recorded every 2 ms before and after the application of the sOMS model (see Table~\ref{table:eventssupp}).

\begin{table}[ht]
\centering
\begin{tabular}{|c|c|c|c|c|}
\hline
\textbf{Trials} & \textbf{\# Events} & \textbf{\# sOMS Events}  & \textbf{Percentage}\\ \hline
10               & 14747.03 $\pm$ 2029.17                        & 2189.15 $\pm$ 385.57     & 85.16\%   \\ \hline
\end{tabular}
\caption{Mean and standard deviation of the events count incoming from the DVS and outgoing from the sOMS model, and the mean number of suppressed events across 10 trials.}
\label{table:eventssupp}
\end{table}

The trials were performed by alternating phases of slightly shaking the DVS sensor while moving a hand in front of the camera, with intermittent pauses.
The sOMS significantly reduced the number of incoming events ($\sim$85.16\%) from the static background, thereby decreasing the number of events to be processed (see Figure~\ref{fig:ev_supp}).

\subsection{sOMS model: EVIMO Benchmark}

To validate the ability of the sOMS model (as described in Section~\ref{ch:soms}) to enhance and segment object motion, the model's response was benchmarked using the EVIMO dataset~\cite{mitrokhin2019ev}.
The EVIMO dataset~\cite{mitrokhin2019ev} is a motion segmentation dataset designed for event cameras. It provides high-resolution temporal data for egomotion estimation and motion segmentation, featuring pixel-wise object masks and 6-DoF poses for both the camera and the moving objects.


\begin{figure*}[!ht]
    \centering
\includegraphics[width=1\textwidth]{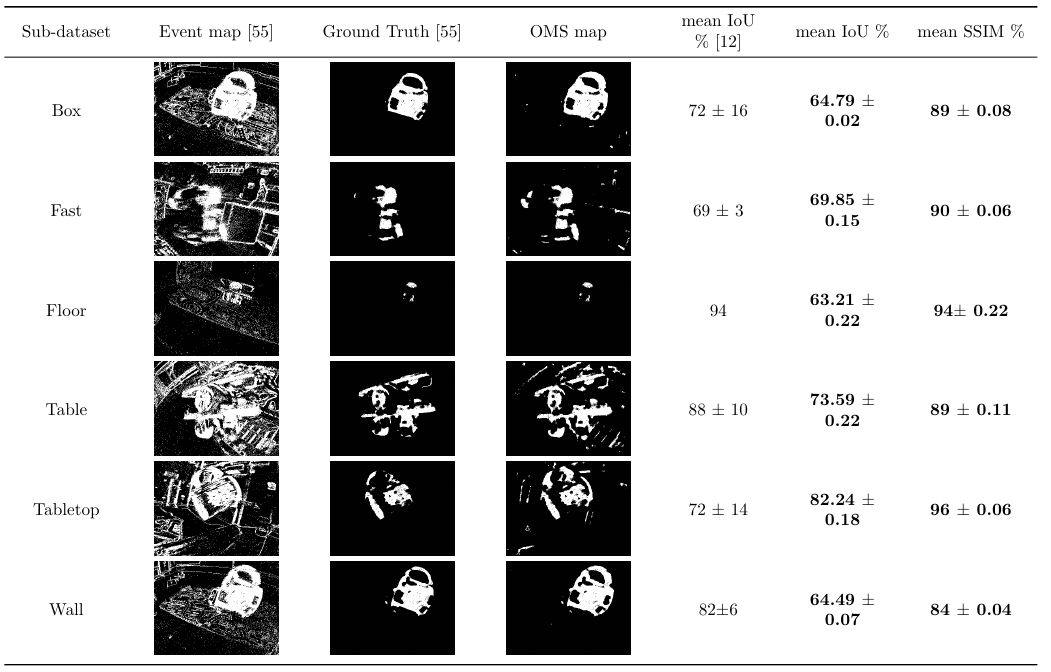}
\caption{From left to right: Sub-dataset name, Event map, Ground truth binary mask, OMS map, mean IoU percentage and mean SSIM across the six EVIMO~\cite{mitrokhin2019ev} sub-datasets. The figure presents results from~\cite{clerico2024retina} alongside our own for comparison.}
\label{fig:EVIMO}
\end{figure*}


The sOMS model for the benchmark employs $\alpha = 0.80$ and $\tau = 0.02$ to align with the EVIMO dataset's Event map and mask collection.
In Figure~\ref{fig:EVIMO}, the model is qualitatively and quantitatively assessed by comparing the pixel-wise object masks with the OMS maps across six different sub-datasets: \textit{Box}, \textit{Fast}, \textit{Floor}, \textit{Table}, \textit{Tabletop}, and \textit{Wall}. Each sub-dataset contains different sequences of motion: \textit{Box} (6), \textit{Fast} (3), \textit{Floor} (2), \textit{Table} (4), \textit{Tabletop} (4), and \textit{Wall} (2).
To quantitatively assess the sOMS model, the mean IoU percentage (as described in~\cite{clerico2024retina}) and the SSIM were computed between the ground truth mask and the OMS map (Equation~\ref{eq:IoUSSIM}).

\begin{equation}\label{eq:IoUSSIM}
\begin{aligned}
\text{IoU} &= \frac{|A \cap B|}{|A \cup B|} \\
\text{SSIM}(x, y) &= \frac{(2\mu_x \mu_y + C_1)(2\sigma_{xy} + C_2)}{(\mu_x^2 + \mu_y^2 + C_1)(\sigma_x^2 + \sigma_y^2 + C_2)}
\end{aligned}
\end{equation}

Figure~\ref{fig:EVIMO} presents the benchmark results against~\cite{clerico2024retina}, demonstrating comparable performance in terms of IoU percentage for the \textit{Fast} dataset, lower performance for the \textit{Box}, \textit{Table}, and \textit{Wall} datasets, and improved performance for the \textit{Tabletop} dataset. The SSIM results indicate a superior and more consistent performance of the model compared to the insights provided by the IoU values.
The observed difference in performance between SSIM and IoU can be attributed to the fact that SSIM focuses on capturing structural similarities and subtle texture variations, making it more forgiving of small discrepancies. In contrast, IoU is more rigid, disproportionately penalizing even minor misalignments or boundary errors in object overlap. Furthermore, the results from the sOMS model are slightly lower than those of the OMS model by Clerico \emph{et al.}~\cite{clerico2024retina}, which can be attributed to the sparse response characteristics of the spiking implementation.

\subsection{SNN Proto-object model: Need for proto-object saliency}\label{ch:SNN-Proto}

The OMS map provides motion contrast information agnostic to whether the detected most salient point originates from an actual moving object. To address this limitation, the architecture feeds the OMS model into the SNN Proto-object model, ensuring that the system generates a saliency map highlighting salient proto-objects. 

\begin{figure*}[!ht]
    \centering
\includegraphics[width=.8\textwidth]{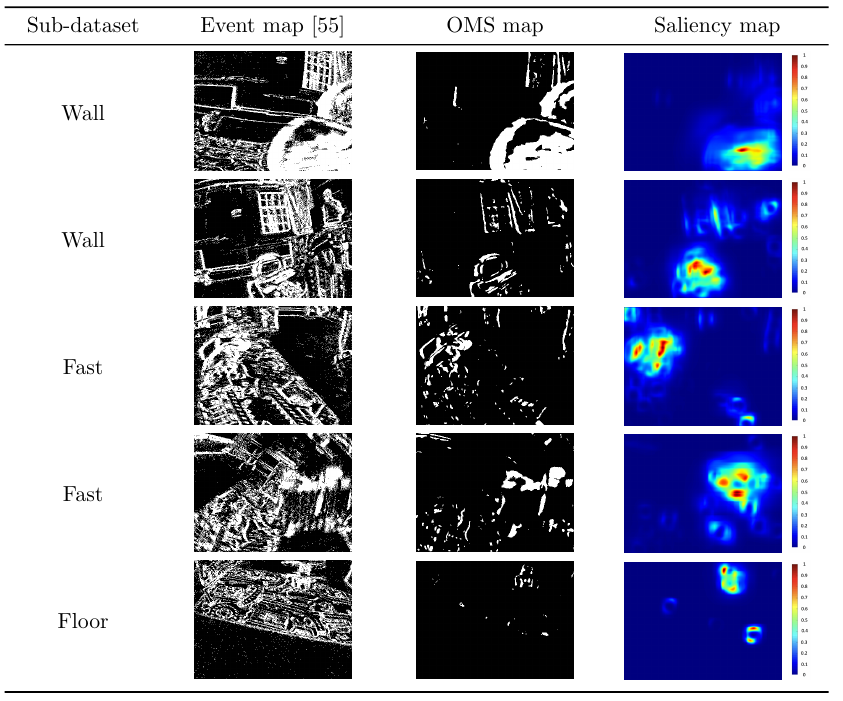}
\caption{From left to right: sub-dataset name, Event map, OMS map and saliency map for several representative examples from the EVIMO dataset~\cite{mitrokhin2019ev}. The colour legend for the saliency map follows the jet map, with blue and red indicating minimum and maximum values, respectively.}
\label{fig:NeedProto}
\end{figure*}


Figure~\ref{fig:NeedProto} presents examples from the EVIMO dataset, illustrating how the saliency map focuses on the object in the scene, whereas the OMS map also highlights parts of the static background, likely due to reflections and rapid acceleration.

To quantitatively assess the accuracy of the SNN Proto-object model in detecting objects, we calculated its accuracy by evaluating whether the coordinates corresponding to the maximum value in the saliency map lie within the ground truth object masks from the EVIMO dataset (Equation~\ref{eq:acc}).

\begin{equation}\label{eq:acc}
\text{Accuracy} = \frac{\sum_{t=1}^{T} \delta\left(\mathbf{B}(t) \cap \mathbf{M}(t) \neq \emptyset\right)}{T} \times 100
\end{equation}

Where \(\mathbf{x}_{\text{max}}(t)\) represents the coordinates of the maximum value in the saliency map at time step \(t\), \(\mathbf{B}(t)\) is the bounding box created around \(\mathbf{x}_{\text{max}}(t)\) (8$\times$8 pixels), \(\mathbf{M}(t)\) is the object mask at time step \(t\), and \(\delta(\cdot)\) is an indicator function that returns 1 if the bounding box \(\mathbf{B}(t)\) intersects with the mask \(\mathbf{M}(t)\). Given the input resolution (346$\times$260 pixels), the bounding box $\mathbf{B}(t)$ is empirically defined as an 8$\times$8 pixel window (approximately 0.07\% of the input) centered on $\mathbf{x}_{\text{max}}(t)$, selected to balance spatial tolerance for saliency shifts with precision, while avoiding the inclusion of multiple objects or irrelevant background. Table~\ref{table:accuracy} shows the mean accuracy percentages and standard deviations for each sub-dataset.

\begin{table}[ht]
\centering
\setlength{\tabcolsep}{12pt} 
\begin{tabular}{|c|c|c|c|c|c|}
\hline
\textbf{Box} & \textbf{Fast} & \textbf{Floor} & \textbf{Table} & \textbf{Tabletop} & \textbf{Wall} \\ \hline
73.4\% $\pm$ 21.61 & 70.9\% $\pm$ 17.85 & 81\% $\pm$ 5.88 & 68.1\% $\pm$ 23.57 & 87.9\% $\pm$ 16.77 & 88.8\% $\pm$ 5.32\\ \hline
\end{tabular}
\caption{Accuracy of the SNN Proto-object model (in percent) for object detection is reported for each EVIMO sub-dataset~\cite{mitrokhin2019ev}.}
\label{table:accuracy}
\end{table}

The SNN Proto-object model demonstrates to effectively detects the salient object in motion in all cases with a slower performance for the \textit{Table}.
The high standard deviations for \textit{Box}, \textit{Fast}, \textit{Table}, and \textit{Tabletop} can be attributed to rapid camera motion, which leads to poor performance by the sOMS model, directly impacting object detection.
Once again, this provides insight that the model would perform best in the case of small adjustment movements.

\subsection{sOMS \& SNN Proto-object model: LLE-VOS Benchmark, challenging light conditions}

To evaluate the system's performance across a broader range of scenarios beyond the EVIMO dataset, which is confined to an office environment, the model has also been benchmarked against the Low-Light Event Video Object Segmentation (LLE-VOS) dataset~\cite{li2024event}. The LLE-VOS dataset comprises 70 video sequences, each containing both standard and low-light versions, along with segmentation labels and low-light event data. These sequences were recorded in diverse environments, including sports facilities, outdoor play areas, educational spaces, conference rooms, and wildlife enclosures, with some sequences involving camera motion and others without. The data are collected using two identical cameras with different exposure times to generate normal-light and low-light pairs of synthetically generated events. Notably, the events used in this benchmark are exclusively derived from low-light conditions, emphasizing the superior capability of event-driven cameras to perform effectively in such challenging environments, where traditional frame-based cameras often face significant limitations.

The dataset was chosen for its variety in both lighting conditions and environments, offering a more comprehensive evaluation of the system's performance in real-world scenarios. Event-based cameras are known for their ability to capture events with high temporal resolution, making them particularly well-suited for challenging lighting situations. By including both normal and low-light images alongside the event data, the dataset enables an exploration of how the system performs under conditions that might pose difficulties for conventional frame-based applications. The system uses the same parameters as those listed in Table~\ref{table:params}.

\begin{figure*}[!ht]
    \centering
\includegraphics[width=1\textwidth]{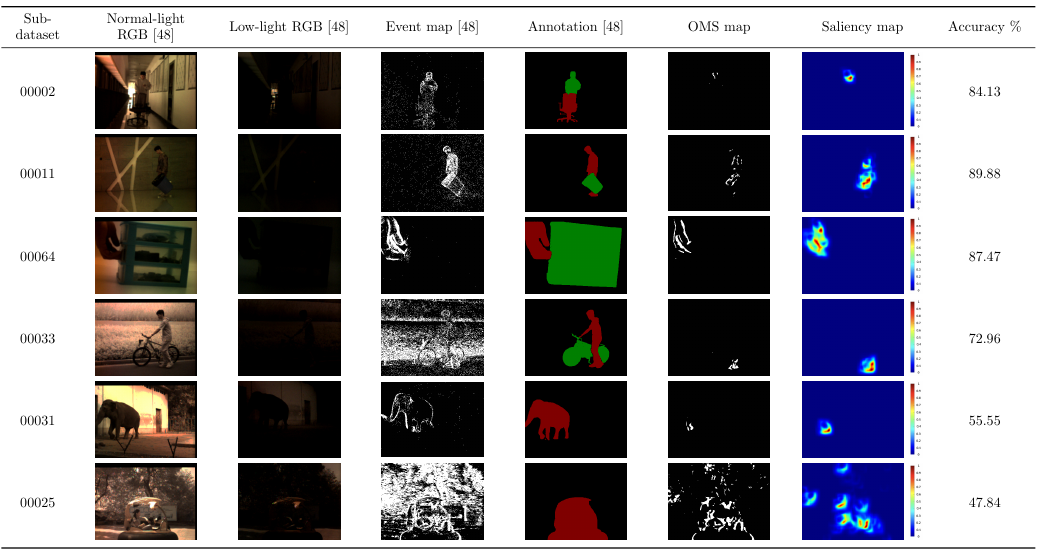}
\caption{From left to right: sub-dataset name, Normal-light RGB frame, Low-light RGB frame, Event map, Annotated frame, OMS map, saliency map and accuracy in percentages for several representative examples from the LLE-VOS dataset~\cite{li2024event}. The colour legend for the saliency map follows the jet map, with blue and red indicating minimum and maximum values, respectively.}
\label{fig:lle-vos}
\end{figure*}


Figure~\ref{fig:lle-vos} shows the normal-light RGB frame, low-light RGB frame, event map, annotation, OMS map, saliency map, and accuracy percentages for examples from the LLE-VOS sub-datasets. 
The Figure presents six examples with varying degrees of accuracy to provide a better understanding of the system's behaviour on the data. The accuracy values are computed as described in Section~\ref{ch:SNN-Proto}.
The system is able to segment motion in indoor and outdoor conditions (i.e. sub-dataset 11 and 33), reaching up to 89.88\% accuracy, generating an OMS map and saliency map coherent with the nature of the event-based data. Example 25 demonstrates the ability of the SNN Proto-object model to aggregate the response to moving objects while disregarding clutter (e.g., foliage). However, similar to Example 31, and coherently with the results on speed found in the characterisation of the system in Section~\ref{ch:charOMS}, the camera motion in Example 25 is too rapid, causing the system to fail in accurately segmenting moving objects, resulting in accuracy rates of 47.84\% and 55.55\%, respectively.
The final mean accuracy in detecting objects across all sub-datasets is 53.3\%, with a standard deviation of 25.56, reflecting the large variability in the fast and slow camera motion sub-datasets. Although the system focuses on the accuracy detection of a single salient object, its performance is slightly lower than that reported in \cite{li2024event} (67.8\%), which employs a learning approach to object detection. This underscores the potential of the sOMS and SNN Proto-object model for future implementations.


\begin{figure*}[ht!]
    \centering
    \includegraphics[width=1\textwidth]{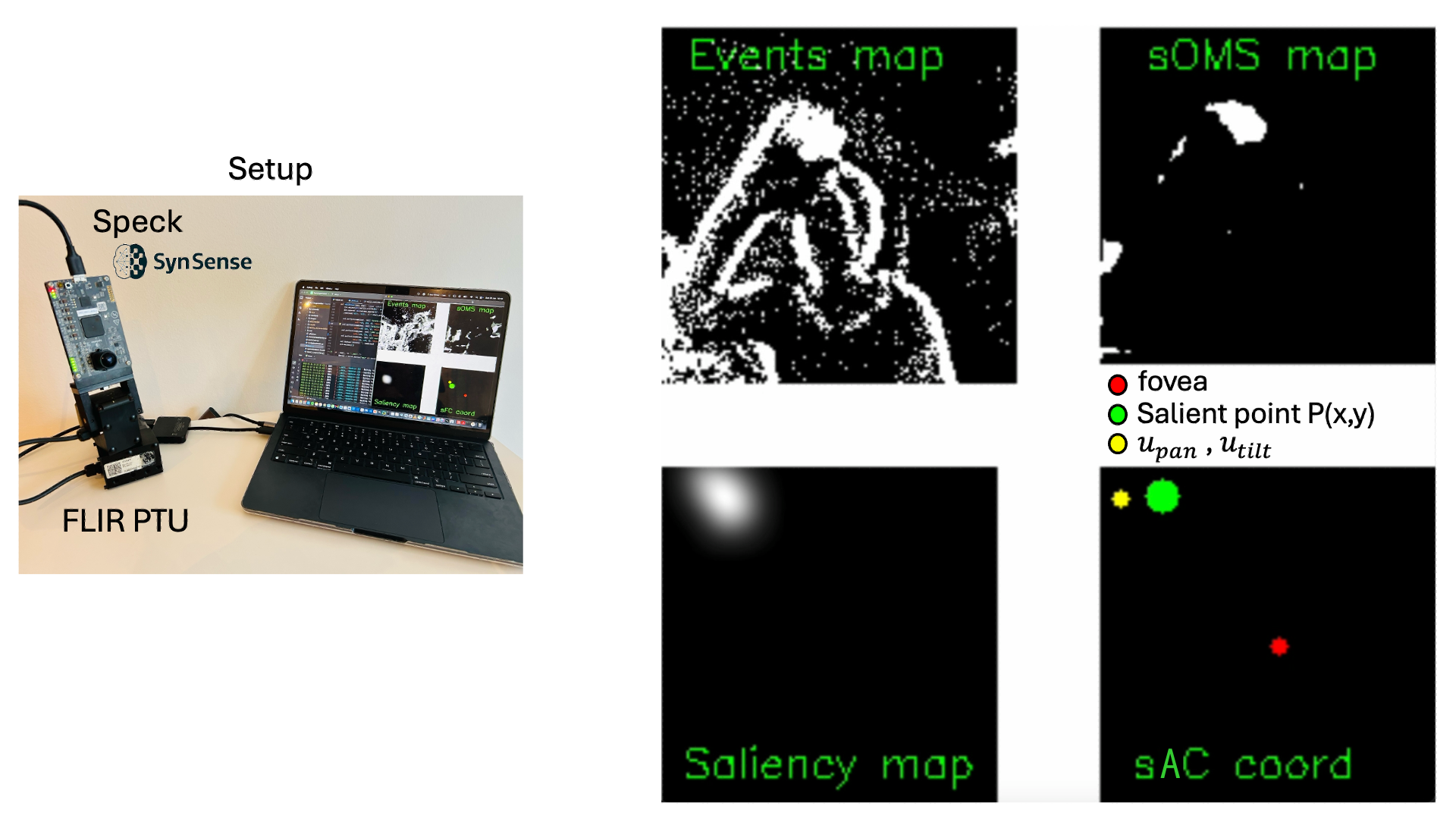}
    \caption{Real-world demonstrator, the Bioinspired attention system. On the left, a view of the setup with the Speck device from Synsense mounted on a FLIR PTU, and the computer displaying the response view. On the right, the response view is shown, including the Events map, the sOMS map, sAC coordinates (where the red dot is the fovea, the green dot is the salient point $P(x,y)$ and the yellow dot sAC commands ( $u_{pan}$ \& $u_{tilt}$).}
    \label{fig:demo}
\end{figure*}


\begin{figure*}[ht!]
    \centering

\includegraphics[width=1\textwidth]{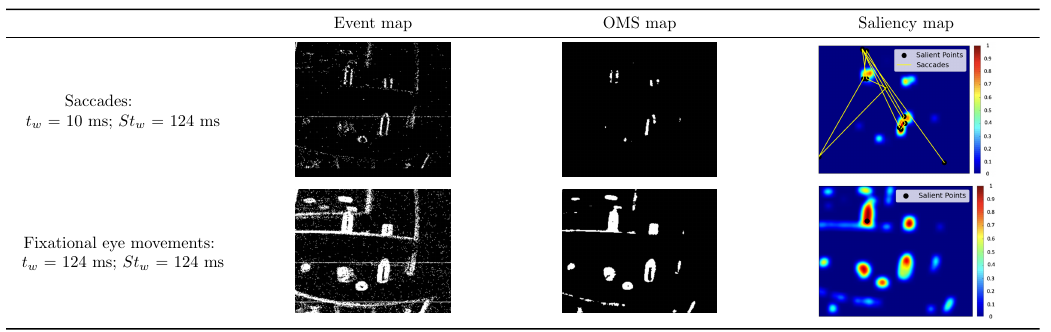}
\caption{From left to right: time window ($t_{w}$) and Saliency map time window ($St_{w}$, Event map, OMS map and Saliency map at $t_{w}$. Black dots on the saliency map represent the salient points, and the yellow lines the saccades trajectories.}
\label{fig:scanpath}
\end{figure*}


\subsection{Real-world demonstration: Bioinspired attention system wandering around}

The Bioinspired attention system (see Figure~\ref{fig:demo}) was demonstrated in various environments to evaluate its robustness and performance. These environments included: (1) an indoor scenario with alternate hand-shaking to test stability, (2) an indoor scene with a person and animals moving dynamically, (3) an outdoor scene with challenging sunlight conditions, and (4) an indoor scene where a person was positioned at a distance from the sensor and waved at the camera. Across these scenarios, the system exhibited a mean latency of \(0.124\) seconds to perform the object motion segmentation and object detection, generating the salient point, demonstrating a rapid response to saccadic commands and bringing the desired salient point to the fovea. 
Specifically, $0.0042 \pm 0.0036$ seconds (over 10 trials) for the object motion sensitivity module and the proto-object detection module, and $0.12 \pm 0.06$ seconds for the visual attention module.
The introduction of fixational eye movements, which are small adjustment movements compared to fast saccadic movements, is essential for performing object motion segmentation accurately and successfully detecting the next salient point of interest.
The video showing the real-time demonstration can be found here: \href{https://youtu.be/dcAJlDgVR0o}{https://youtu.be/dcAJlDgVR0o}.

\subsubsection{Bioinspired attention system wandering around: evaluating the scanpath}
To further validate the proposed architecture in robotic scenarios and underscore the necessity of spatiotemporal integration, we evaluated the system in a static setting involving objects placed on a shelf. Events were generated using the ATIS cameras mounted on the iCub robot, which executed eye movements through the motorised actuation of the ``eye muscles,'' thereby mimicking fixational eye movements. The eye movements covered 72 discrete positions along a circular trajectory, with updates occurring every 5~ms, resulting in a total duration of 360~ms and a pan-tilt angle of $1^\circ$ per position. During this process, we recorded the system’s output, including the maximum salient points, saliency maps, and simulated saccades (representing the scanpath, namely the collection of eye trajectories).
We considered two distinct operational conditions. In the first case, \textit{Saccades}, we assumed no time for spatiotemporal integration. The system computed the Event map, OMS map, and Saliency map at every time window $t_{w} = 10$~ms, simulated the saccade, and subsequently generated the final cumulative saliency map using a saliency map time window $St_{w}$. In the second case, \textit{Fixational eye movements}, we set $t_{w} = St_{w}$ = 124~ms (same latency of the real-world demonstrator), allowing the system to integrate information temporally within a single, consistent window.
The responses in Figure~\ref{fig:scanpath} further underscore the need for spatiotemporal integration to enable meaningful object detection (\textit{Fixational eye movements} case) and prevent random saccadic jumps to non-informative locations (\textit{Saccades} case), while also demonstrating the model’s effectiveness in static scenarios. This analysis lays the foundation for future research on robotic eye movements for event-based active vision.

\section*{Code and Data Availability}

\textcolor{black}{
The code is available at \url{https://github.com/GiuliaDAngelo/Speckegomotion}. \\
The code for the demonstration is available at \url{https://github.com/GiuliaDAngelo/Speckegomotion/tree/demo}. \\
The code to reproduce the data for the characterisation is available at \url{https://github.com/GiuliaDAngelo/Speckegomotion/blob/main/CreateStimuli.py}. \\
All the data used in the study can be easily reproduced by running the scripts provided in our GitHub repository. Furthermore, the two benchmark datasets employed are already publicly accessible online (EVIMO~\cite{mitrokhin2019ev} \& LLE-VOS~\cite{li2024event}).}

\section{Discussion}

The characterisation of the sOMS model confirmed the sigma sizes proposed in~\cite{clerico2024retina}. However, the model exhibits a comparable mean MFR and mean ISI even with $\sigma_{c}=2$. Additionally, the mean MFR indicates increased activity for a larger kernel size ($s=32$), while maintaining a similar mean ISI. This suggests comparable performance and allows us to exclude the use of larger kernel sizes, as they result in higher MFR. The system performs qualitatively well across various scenarios with diverse spatial frequencies, confirming the model's scale invariance. The characterisation also demonstrates that the model does not completely suppress activity in the \textit{Eye Only} condition but effectively performs object motion segmentation, enhancing motion contrast in all \textit{Eye+Object} cases where the foreground speed is greater than the background speed ($s_{F} > s_{B}$). 

The sOMS model effectively suppresses 85\% of events while achieving a mean IoU of up to 82.24\% and a mean SSIM accuracy of 96\% in multi-object motion segmentation compared to the ground truth, demonstrating that the majority of the suppressed events are not related to objects in motion.
In addition, although the benchmark comparison of the mean IoU (\%) indicates lower performance compared to~\cite{clerico2024retina}, likely due to the sparse responses generated by the sCNN layer, our results demonstrate greater consistency across the entire dataset as demonstrated by the standard deviation of both mean IoU and mean SSIM.
The model also demonstrates its effectiveness across diverse scenarios, as evidenced by the characterisation with grating stimuli and benchmarking across all the EVIMO sub-datasets.
Akin to the event-based nature of the input, the object motion segmentation in the sOMS map generates a sparse mask that highlights the leading and trailing edges of objects moving in the scene.
The sOMS model demonstrates consistent performance even with the LLE-VOS benchmark, which contains a variety of dynamic indoor and outdoor scenes.
As expected from the sOMS characterisation, the model fails when the camera motion exceeds the speed of the dynamic entities in the scene. This further reinforces the idea that the model performs best with small sensor movement adjustments before detecting the salient object and performing a saccade toward the next one.

The introduction of the SNN Proto-object model adds an additional processing layer to detect potential objects in the scene, further refining object motion segmentation and perceptually grouping the visual scene information. The SNN Proto-object implementation, adapted on the \textit{sinabs} library from the previously proposed attention model~\cite{Dangelo_etal22}, confirms its ability to accurately detect objects in the scene.
The SNN Proto-object model, grounded in the principles of the Gestalt laws, facilitates the perceptual grouping of edge features with close contours, effectively disregarding lines and potential clutter.
As shown in Figure~\ref{fig:attchar}, the SNN Proto-object model enhances the sOMS model's ability to detect potential objects in the scene, achieving an accuracy up to 88.8\% (as shown in Figure~\ref{fig:NeedProto}). 
Additionally, the sOMS generates the OMS map, producing thicker edge maps of moving objects and thereby facilitating proto-object detection.
Consistent with the event-based nature of the sOMS input, the saliency map detects moving objects with 89.88\% accuracy on the LLE-VOS dataset under challenging lighting conditions, effectively focusing attention on the most salient point in the scene.
As shown by the high standard deviation across the entire dataset (25.56), integrating the sOMS model with the SNN Proto-object model is most effective in setups where movements consist of small adjustments to detect the maximum salient point and foveate toward it.
From an active vision perspective, this aligns with the non-uniform distribution of the retinal structure, which concentrates high resolution in the center of the visual field—the fovea—to facilitate further object recognition.

Both learning-free models demonstrate robust performance across a wide range of scenarios, effectively handling motion segmentation and detection without the need for computationally expensive training or adaptation to varying conditions. This opens the possibility of integrating learning algorithms for more complex tasks, such as object recognition.

Thanks to the sAC control, the complete closed-loop real-time bioinspired attention system performs effectively, demonstrating robustness across various scenes. The additional fixational eye movements following the previous saccade (at the maximum salient point) facilitate precise adjustment movements, enabling reliable and consistent object motion segmentation in local motion. This suggests that the system detects objects in this context, primarily using saccades to transition from one salient point to the next. As indicated by Crevecoeur et al.~\cite{crevecoeur2017saccadic}, the partial saccadic suppression may result from efficient sensorimotor computations. This aligns with our findings, demonstrating that the bioinspired attention system can accurately segment and detect object motion only when these small adjustment movements are incorporated, allowing saccades to be used for transitioning between salient points.

\section{Conclusion}

The proposed end-to-end bioinspired attention system demonstrates robust performance in segmenting object motion during fixational eye movements and saccades toward the most salient proto-object. Detailed characterisation and benchmarking of the sOMS and SNN Proto-object models show that the system enhances motion contrast, prioritises coherent objects, and disregards background features across varied scenarios, including office environments and low-light indoor/outdoor scenes.

This architecture introduces an innovative visual attention system through object motion segmentation that relies exclusively on visual cues, maintaining high temporal resolution without requiring calibration or synchronization with external sensors such as IMUs. This design aligns with the absence of proprioceptive capabilities in retinal ganglion cells (RGCs), as there is no scientific evidence to suggest that RGCs possess proprioceptive functions. Consequently, proprioceptive feedback from external devices is not necessary for effective object motion segmentation during fixational eye movements.
Furthermore, the saliency-based attention model can perceptually group motion features, enabling the detection of potential salient objects coherently with the nature of the even-based input.
The resulting attention mechanism, based on object motion segmentation, reveals a single salient point, requiring a shift of the gaze to place it in the fovea for further stable inspection with high resolution.
Implementing the non-uniform structure of the retina, which is beyond the scope of this work, would enhance the consistency of the proposed system. As demonstrated by D'Angelo \emph{et al}.~\cite{d2020event}, this would further reduce noise, decrease the number of retinal events, and enhance motion direction detection through the varying receptive field sizes from the periphery to the fovea, offering deeper insights into human selective attention. This highlights a crucial consideration regarding camera resolution: a high resolution is likely, not necessary, as we can actively direct the gaze toward the point of interest.

From an active vision perspective, the central finding of this work is the need for small corrective movements, such as fixational eye movements, for spatiotemporal integration. These movements effectively highlight object motion, enabling the stable detection of salient points, which is crucial for attentional shifts~\cite{gu2024microsaccades}, while reserving saccades solely for transitioning to the next salient point. Without fixational eye movements, the system would shift abruptly from one salient point to another, failing to stabilise the visual field and hindering the detection of stable points of interest.
The partial suppression of information during saccades, as demonstrated in our characterisation of the sOMS model, aligns with recent literature suggesting that information processing continues even during saccades, while perceptual stability is maintained. Specifically, it has been shown that saccades suppress visual processing of both color and luminance in the early visual cortex, but do not completely halt it~\cite{zhang2024execution}.
Furthermore, presaccadic visual content can modulate postsaccadic processing, further indicating ongoing visual processing during eye movements~\cite{stankov2021during}.
These findings suggest that our visual system partially maintains continuous processing during saccades while preserving perceptual stability and preventing disorientation from constant visual motion.

This underscores the embodied nature of our system, where active perception, through the interplay of fixational eye movements and saccades, directly shapes visual processing, mirroring biological mechanisms that integrate motor actions with sensory input to enhance stability and perception. Central to future experiments will be the investigation of saccades and fixational eye speeds, exploring the limits of the system and the PTU to achieve smoother saccadic movements and even smooth pursuit. Moreover, even if the scene remains static, the system is able to \textit{wander around} effectively, thanks to the fixational eye movements and attention mechanisms that generate a bottom-up saliency map of the scene. Additionally, small retinal shifts create depth-dependent motion patterns for objects at different distances, leading to spike variations that provide an initial cue for monocular depth perception.
A further development could integrate recurrence between the border ownership layer and the grouping layer enhancing a more dense segmentation of the objects~\cite{d2024event}.

The entire architecture is built using spiking neural networks, representing the first closed-loop neuromorphic visual attention approach through object motion sensitivity. The decision to implement the visual and control components on separate neuromorphic frameworks is based on the fact that at the time of writing the entire pipeline could likely not be accommodated on a single platform. This approach allows for the distribution of the architectures across different hardware leveraging the neuromorphic intermediate representation~\cite{pedersen2024neuromorphic}.
The full deployment of the architecture is beyond the scope of this work; however, this study serves as the first attempt to characterise end-to-end active visual attention and lays the foundation for its complete implementation on neuromorphic hardware to demonstrate the advantages of end-to-end neuromorphic systems in terms of reduced latency and power consumption.
The architecture is entirely model-based and not dependent on training, allowing the system to handle diverse environments and scenarios, as demonstrated in various experiments, ranging from ideal gratings to office scenarios and challenging indoor and outdoor environments. However, learning could be employed to adapt the parameters to further optimise the system's performance.

\section{Acknowledgement}

G.D. acknowledges the financial support from the European Union’s HORIZON-MSCA-2023-PF-01-01 research and innovation programme under the Marie Skłodowska-Curie grant agreement ENDEAVOR No 101149664. M.H. was supported by the European Union under the project ROBOPROX (reg. no. CZ.02.01.01/00/22\_008/0004590). 
We thank Bedrich Himmel for assistance with the PTU and modelling the 3D printed support for the Speck device and Suman Ghosh for the insightful and inspiring discussions. 
We also thank the Telluride Engineering Neuromorphic Workshop for providing the opportunity to connect and explore this project.

\section*{Author contributions statement}

G.D. and A.H. conceived the main idea behind this work. G.D., A.H., and M.F. developed the code for the entire architecture. Specifically, A.H. focused on the real-time demonstrator, M.F. on Spiking Attention Control, and G.D. on the entire architecture.
The experiments were conceived by G.D., with the constant support of A.H., M.F., and V.C., with supervision from M.H. 
G.D. performed and analysed the experiments with assistance from V.C., A.H., and M.F., under the supervision of C.B. and M.H. 
G.D. wrote the manuscript with contributions from A.H., M.F., and V.C., under the supervision of C.B. and M.H.




\end{document}